%% file: main.tex
\documentclass[10pt,twocolumn,letterpaper]{article}

\usepackage[pagenumbers]{iccv} 


\usepackage{multirow}
\usepackage{float}
\usepackage[T1]{fontenc}
\usepackage{makecell}
\definecolor{iccvblue}{rgb}{0.21,0.49,0.74}
\usepackage[pagebackref,breaklinks,colorlinks,allcolors=iccvblue]{hyperref}

\usepackage{listings}
\usepackage{xcolor}
\usepackage{hyperref}

\def\paperID{13389} 
\def\confName{ICCV}
\def\confYear{2025}

\title{Simulating Dual-Pixel Images From Ray Tracing For Depth Estimation}

\author{
Fengchen He, Dayang Zhao, Hao Xu, Tingwei Quan, Shaoqun Zeng\\
Huazhong University of Science and Technology \\
{\tt\small \{linyark, dayangzhao, xuhao\_2003, quantingwei, sqzeng\}@hust.edu.cn}
}

\begin{document}
\setlength{\abovedisplayskip}{2mm}
\setlength{\belowdisplayskip}{2mm}

\maketitle
\input{sec/0_abstract}

\input{sec/1_intro}
\input{sec/2_related}

\input{sec/3_method}

\input{sec/4_experiment}

\input{sec/5_conclu.tex}

{
    \small
    \bibliographystyle{ieeenat_fullname}
    \bibliography{main}
}

\input{sec/X_suppl}

\end{document}

%% file: sec/0_abstract.tex
\begin{abstract}

Many studies utilize dual-pixel (DP) sensor phase information for various applications, such as depth estimation and deblurring.
However, since DP image features are entirely determined by the camera hardware, 
DP-depth paired datasets are very scarce, especially when performing depth estimation on customized cameras.
To overcome this, studies simulate DP images using ideal optical models.
However, these simulations often violate real optical propagation laws,
leading to poor generalization to real DP data.
To address this, we investigate the domain gap between simulated and real DP data, 
and propose solutions using the Simulating DP Images from Ray Tracing (Sdirt) scheme.
Sdirt generates realistic DP images via ray tracing and integrates them into the depth estimation training pipeline.
Experimental results show that models trained with Sdirt-simulated images
generalize better to real DP data.
The code and collected datasets will be available at \href{https://github.com/LinYark/Sdirt}{https://github.com/LinYark/Sdirt}.

\end{abstract}

%% file: sec/1_intro.tex
\section{Introduction}
\label{sec:intro}

\begin{figure}[h]
  \centering
  \includegraphics[width=1\linewidth]{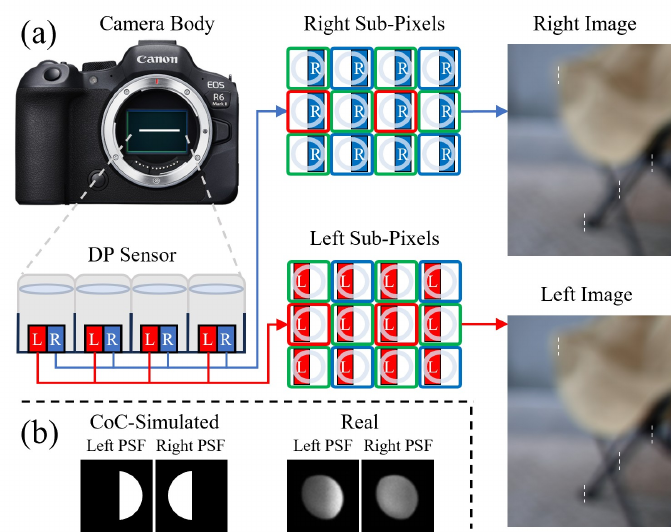}
  \vspace{-7mm}
  \caption{
    \label{fig:i1}  
    (a) Imaging process of a DP camera. The slight shifts between the left and right DP images caused by phase differences are illustrated with white dashed lines.  
    (b) Example comparison between real and CoC-simulated DP PSFs, showing a significant difference between them.
  }
\end{figure}

The DP sensor~\cite{kobayashi2016low,shim2021all} is designed to split each pixel into left and right sub-pixels, 
utilizing microlenses and sub-pixels to achieve phase splitting.
This allows the capture of a pair of images in one shot, known as DP images, as shown in \cref{fig:i1}(a).  
DP images are not only used for autofocus~\cite{sliwinski2013simple,ho2020af,choi2023exploring},  
but also improve performance in tasks such as deblurring~\cite{yang2023k3dn,li2022learning,abuolaim2022improving} and depth estimation~\cite{kang2022facial,garg2019learning,pan2021dual}.  
In the field of depth-from-dual-pixel (DfDP) estimation, the scarcity of DP-depth paired datasets stems from hardware limitations.  
To address this, many studies~\cite{xin2021defocus,punnappurath2020modeling,li2024dual,pan2021dual,abuolaim2021learning,li2023learning} have focused on simulating DP images using RGBD datasets,  
which serve as more flexible and accessible data sources.

The key to simulating DP images from RGBD datasets lies in accurately modeling the DP point spread function (PSF).
In recent years, numerous model-based and calibration-based DP PSF simulators have been proposed.
Calibration-based simulators~\cite{xin2021defocus,li2023learning,li2024dual} require substantial time calibrating real cameras to match DP PSFs,
but suffer from issues such as interpolation errors due to discrete calibration points and the neglect of lens and DP sensor parameters, limiting model transferability.
Model-based simulators~\cite{punnappurath2020modeling,pan2021dual,abuolaim2021learning} employ ideal optical or aberration analysis models to directly compute DP PSFs, bypassing calibration.
However, these models often perform poorly on real DP images because they ignore lens aberrations and sensor phase splitting characteristics.
As shown in \cref{fig:i1}(b), the ideal thin-lens model's simulated circle-of-confusion (CoC) DP PSF exhibits a significant domain gap with the real PSF, adversely affecting depth estimation.

To bridge the domain gap between real DP images and DP images simulated by model-based simulators, we propose Sdirt.
Sdirt consists of a ray-traced DP PSF simulator and a pixel-wise DP image rendering module, which can accurately generate DP images with aberration and phase information.
Specifically, given a fixed focus lens and DP sensor system with known parameters, 
we use ray tracing~\cite{wang2022differentiable,yang2024curriculum,yang2023aberration} to calculate the spatially varying DP PSFs and train a multilayer perceptron (MLP) to predict them. 
Subsequently, we convolve per-pixel DP PSFs with the all-in-focus (AiF) image to simulate the DP images captured by a real camera.
After training a DfDP model with DP images simulated by Sdirt, the model can accurately estimate depth using optical aberration and phase information, enhancing its ability to generalize to real DP images.

We conduct extensive experiments to validate the effectiveness of Sdirt.
The experimental results demonstrate that, compared to other model-based simulators, 
the DP PSFs and DP images generated by Sdirt better resemble real data, 
and the DfDP model trained on Sdirt achieves superior generalization.
In summary, our contributions are fourfold:
\begin{itemize}
\item We propose a ray-traced DP PSF simulator that computes spatially varying DP PSFs, addressing the domain gap between simulated and real DP PSFs caused by lens aberrations and sensor phase splitting.
\item We propose a pixel-wise DP image rendering module that uses an MLP to predict the DP PSF for each pixel, narrowing the gap between simulated and real DP images.
\item Depth estimation results show that the DfDP model trained on Sdirt generalizes better to real DP images.
\item We collected DP119, a real DP-depth paired test set with an open lens structure and fixed focus, featuring diverse real-world scenes.
\end{itemize}

%% file: sec/2_related.tex
\section{Related Work}
\label{sec:rw}

\subsection{DP Datasets}
Most professional and mobile cameras have DP sensors, but only Canon cameras~\cite{abuolaim2020defocus,punnappurath2019reflection,punnappurath2020modeling} 
and Google Pixel phones~\cite{zhang20202,garg2019learning,wadhwa2018synthetic} offer DP data. 
In the field of depth estimation, supervised tasks rely on DP-depth paired datasets~\cite{garg2019learning,zhang20202,punnappurath2020modeling,abuolaim2020defocus}. 
However, changes in the lens, sensor, focal length, or aperture settings alter the PSFs, making it necessary to re-acquire DP-depth data.
Consequently, many researchers focus on simulating DP-depth datasets using RGBD data to reduce data collection costs and improve flexibility.

\subsection{DP PSF simulators}
Simulating DP images from RGBD datasets relies on DP PSFs, 
with recent progress in calibration-based~\cite{xin2021defocus,li2024dual,li2023learning} and model-based~\cite{punnappurath2020modeling,pan2021dual,abuolaim2021learning} DP PSF simulators.
In calibration-based simulators, Xin \etal~\cite{xin2021defocus} acquired full-space DP PSFs by sampling discrete points in space and interpolating, 
while Li \etal~\cite{li2024dual} acquired them through size scaling using the CoC model. 
Furthermore, Li \etal~\cite{li2023learning} generated DP PSFs using a U-Net, which requires a large amount of DP-depth paired data. 
Calibration-based simulators are time-consuming and prone to interpolation and scaling errors, while model-based simulators circumvent these issues.
Pan \etal~\cite{pan2021dual} simplified the optical system to an ideal thin lens and used a symmetrically divided rectangular aperture to calculate DP PSFs. 
Punnappurath \etal~\cite{punnappurath2020modeling} modeled the DP PSFs using a depletion kernel that is phase symmetric, meaning the left and right PSFs of the same object point are flip symmetric.
Abuolaim \etal~\cite{abuolaim2021learning} used aberration analysis models to calculate DP PSFs, but the resulting patterns still remained phase symmetric.
These studies oversimplify optical propagation by neglecting aberrations and phase splitting, leading to unrealistic domain gaps.

\subsection{Applications driven by DP data}
In recent years, DP data-driven applications such as depth estimation~\cite{kang2022facial,pan2021dual,garg2019learning,zhang20202,kim2023spatio,pan2024weakly,ghanekar2024passive}, 
deblurring~\cite{yang2023k3dn,li2022learning,abuolaim2022improving,yang2024ldp,abuolaim2021ntire,zhang2022dynamic,ruan2024self}, 
and refocusing~\cite{ho2020af,choi2023exploring,swami2023adjust,alzayer2023dc2} have undergone rapid development.
Additionally, Kang \etal~\cite{kang2022facial} utilized DP data for facial scanning, similar to depth estimation tasks.
Wadhwa \etal~\cite{wadhwa2018synthetic} employed DP data for synthetic shallow depth-of-field imaging, a popular recent application in mobile photography.
Shi \etal~\cite{shi2024split} integrated DP hardware with structured diffractive optical elements for complex image encoding and high-precision multimodal reconstruction.
In addition, DP data can be used for disparity estimation~\cite{wu2024disparity,monin2024continuous}, rain removal~\cite{li2024dual}, and reflection removal~\cite{punnappurath2019reflection}.
It is foreseeable that DP data-driven applications will continue to emerge in various new fields.

%% file: sec/3_method.tex
\begin{figure*}[ht]
    \centering
    \includegraphics[width=1\linewidth]{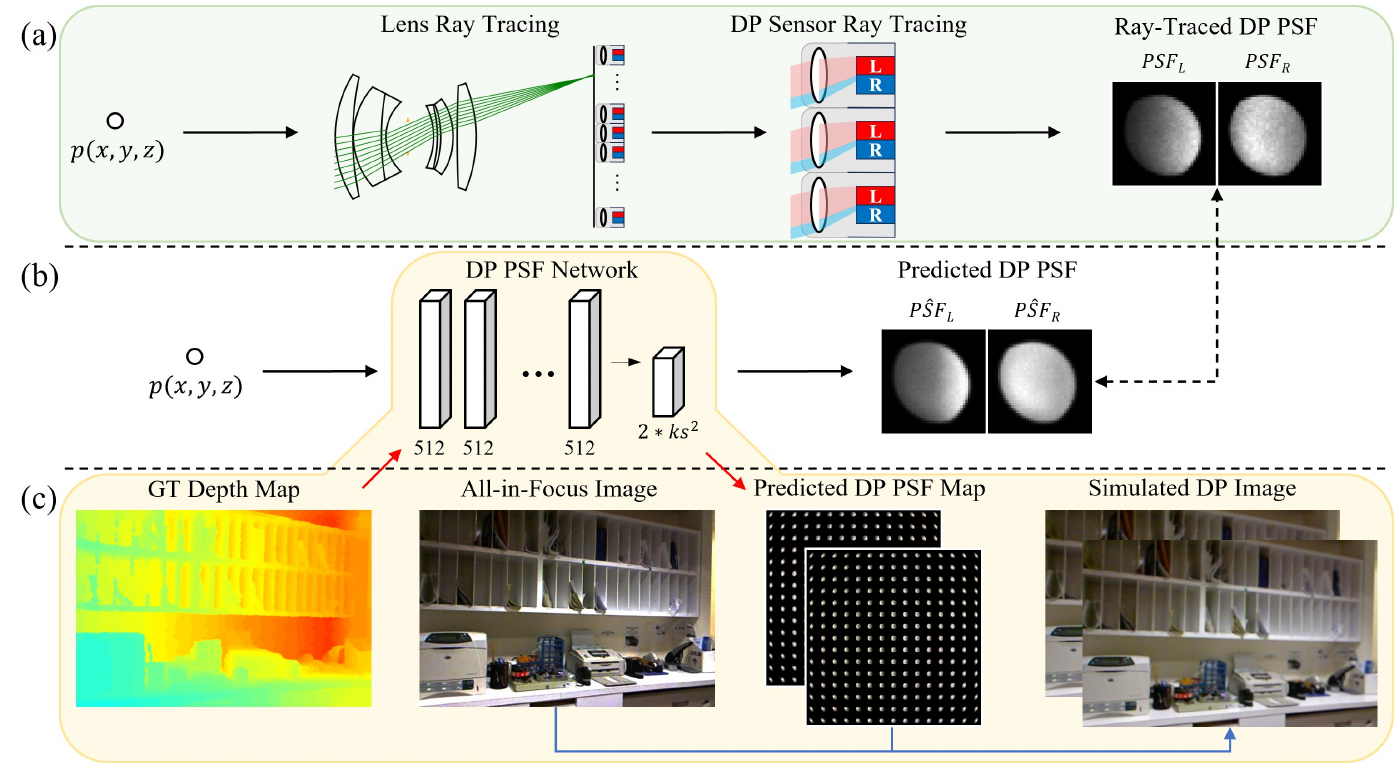}
    \vspace{-6.5mm}
    \caption{\label{fig:m1}\textbf{Simulating Dual-Pixel Images from Ray Tracing pipeline. }
    (a) Ray-traced DP PSF simulator. Calculates spatially varying DP PSFs for lens and DP sensor through ray tracing.
    (b) DP PSF prediction network. Trains an MLP network to predict DP PSFs, using the ray-traced DP PSFs as GT.
    (c) Pixel-wise DP image rendering module. The network predicts the DP PSFs for all points in the depth map (red pass). 
    Then, each DP PSF is convolved with the AiF RGB image to render the simulated DP image (blue pass). 
    }
    \vspace{-2mm}
\end{figure*}

\section{Method of Sdirt}\label{sec:Sdirt}
To provide the required input for training the DfDP network, Sdirt generates realistic DP images through ray tracing.
As shown in \cref{fig:m1}(c), the module achieves pixel-wise rendering of DP images by convolving the DP PSF corresponding to each point in the depth map with the AiF RGB image.
The DP PSF for each point is inferred by a pre-trained MLP network.
As illustrated in \cref{fig:m1}(b), the offline training process of the MLP network uses ray-traced DP PSFs (\cref{fig:m1}(a)) as ground truth (GT).

\begin{figure}[ht]
    \centering
    \includegraphics[width=1\linewidth]{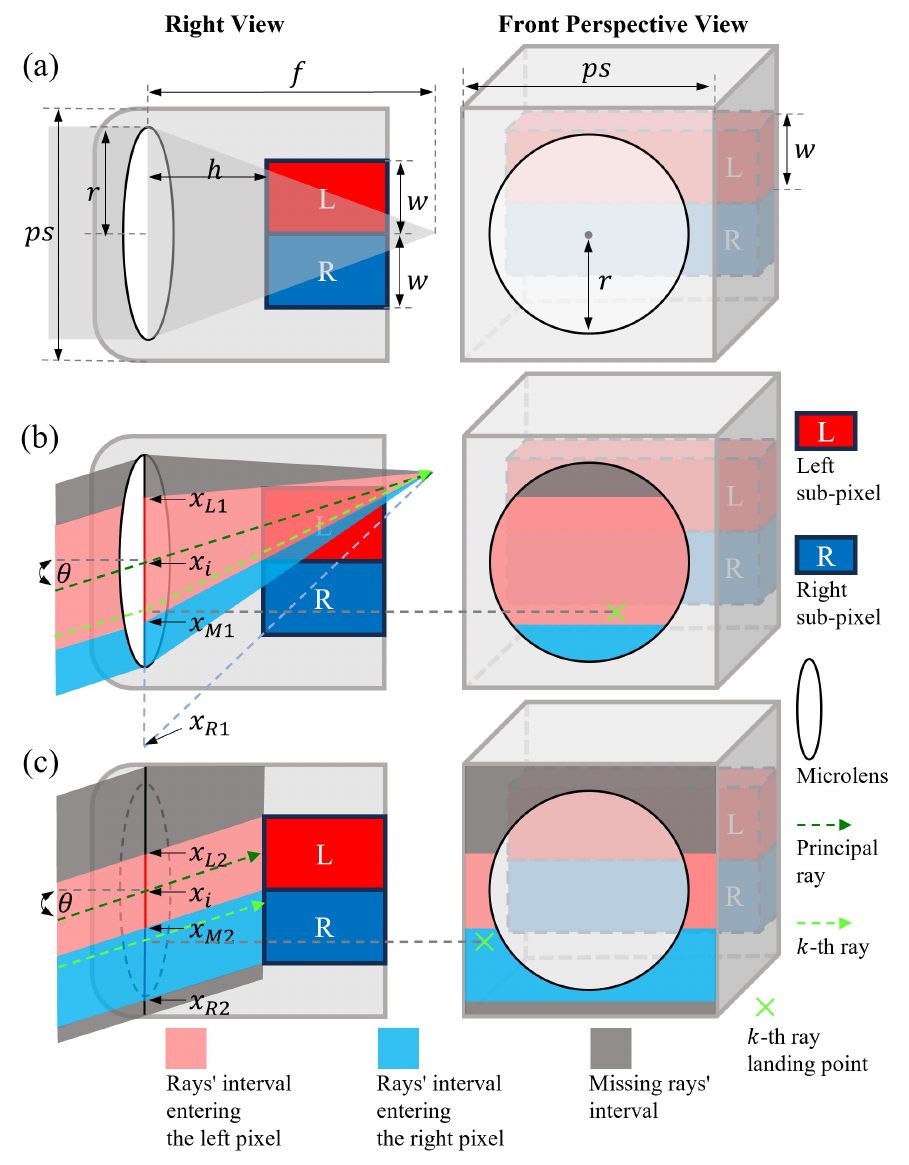}
    \vspace{-8.5mm}
    \caption{\label{fig:m2}
    (a) DP pixel structure layout. The left part is a right-side view, and the right part is a front perspective view.
    (b) On hitting the microlens, the $k$-th ray lands on left/right sub-pixel in red/blue interval, else missed.
    (c) Without hitting the microlens, the $k$-th ray lands on left/right sub-pixel in red/blue interval, else missed.
    }
    \vspace{-3mm}
\end{figure}

\subsection{Ray-traced DP PSF simulator}
As shown in \cref{fig:m1}(a), lens ray tracing can accurately obtain the landing point on the sensor surface~\cite{wang2022differentiable,yang2024curriculum,yang2023aberration,cote2023differentiable}.
We construct a ray set $A$ containing $n$ rays, each originating from an object point $p$.
The aperture stop's image in object space is treated as the entrance pupil, from which $n$ points are densely sampled.
Each ray in $A$ is initially directed from $p$ to its corresponding entrance pupil sample.
As rays in $A$ traverse each lens surface, their positions and directions are updated by refraction according to Snell's law and the lens parameters.
After multiple refractions, the rays reach the sensor plane.
We denote the sensor-plane landing points as $O$ and the corresponding ray directions as $D$.

DP sensor ray tracing is limited because camera manufacturers do not disclose the structural parameters of the microlens and sub-pixel components within the DP pixel.
As a result, we simplify the DP pixel structure (\cref{fig:m2}(a)) based on past research~\cite{kobayashi2016low,shim2021all}. 
We model the microlens as a thin lens with radius $r$ and focal length $f$, 
and define $h$ as distance between the sub-pixel and the microlens, $w$ as sub-pixel width, and $ps$ as DP pixel size.

\begin{figure}
    \centering
    \includegraphics[width=1\linewidth]{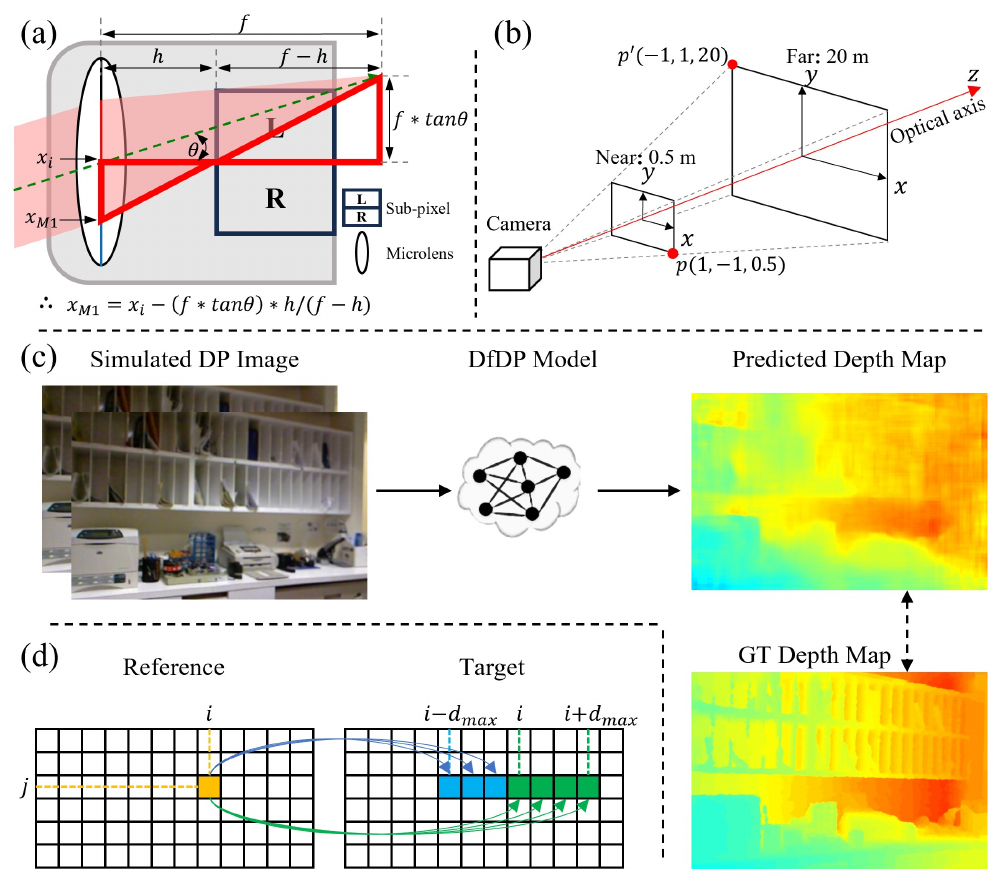}
    \vspace{-7mm}
    \caption{\label{fig:m3} 
    (a) Derivation hints for \cref{eq:f1_0}. 
    (b) The valid imaging region is a frustum, and we normalize the $(x,y)$ coordinates to $[-1,1]$.
    (c) The DfDP model takes DP image as input to predict the depth map.
    (d) During cost volume generation in~\cite{cheng2020hierarchical}, we stack original disparity (green arrows) and add reverse disparity (blue arrows), with $d_{max}$ as max displacement.
    }
    \vspace{-3mm}
\end{figure}

The DP PSF can be obtained by separately calculating the cumulative integration of the rays for each sub-pixel.
We analyze which sub-pixel the $k$-th ray $A_k$ in the ray set $A$ ultimately enters.
Through lens ray tracing, we obtain the landing point $O_k(x_k, y_k, z_k)$ and direction $D_k(x_k', y_k', z_k')$ of the $k$-th ray $A_k$ on the sensor surface.
Based on the landing point $O_k$, we can easily calculate that $A_k$ lies within the DP pixel $(i,j)$, which has surface center coordinates $(x_{i}, y_{i})$.
For $O_k$, the situation differs depending on whether it lies \textbf{within} or \textbf{outside} the microlens:

\noindent\textbf{When $O_k$ lies within the microlens, } as shown in \cref{fig:m2}(b), $A_k$ is refracted by the microlens and then enters a sub-pixel.
According to geometric optics theory, $A_k$ and the principal ray incident on the thin lens at the same angle both converge on the focal plane of the thin lens.
We only need to focus on a few boundary lines $x_{L1}, x_{M1}, x_{R1}$ to determine which sub-pixel $A_k$ ultimately enters.
If $x_k$ is in the interval $[x_{L1}, x_{M1}]$, it enters the left sub-pixel; if in the interval $[x_{M1}, x_{R1}]$, it enters the right sub-pixel; otherwise, it is considered a missing ray.
The calculation methods for the boundary lines are as follows:
\begin{align}
    x_{L1} &= x_i+w - (f*\tan\theta - w)*h/(f-h)  \notag \\
    x_{M1} &= x_i   - (f*\tan\theta)*h/(f-h)      \label{eq:f1_0} \\
    x_{R1} &= x_i-w - (f*\tan\theta +w)*h/(f-h)   \notag
\end{align}
Where $\tan\theta$ is the tangent of the angle at which $A_k$ lands on the sensor surface, equivalent to ${x_k'}/{z_k'}$.
We provide the derivation hints for \cref{eq:f1_0} in \cref{fig:m3}(a). 

\noindent\textbf{When $O_k$ lies outside the microlens, } as shown in \cref{fig:m2}(c), $A_k$ enters the sub-pixel directly without refraction.
Similarly, checking boundary lines $x_{L2}, x_{M2}, x_{R2}$ suffices to determine which sub-pixel $A_k$ ultimately enters:
\begin{align}
    x_{L2} &= x_i+w - h*\tan\theta  \notag \\
    x_{M2} &= x_i   - h*\tan\theta    \label{f1_1} \\
    x_{R2} &= x_i-w - h*\tan\theta  \notag
\end{align}

The left PSF ($PSF_L$) can be calculated by integrating the set of rays $A$ emitted from point $p$ over all left sub-pixels:
\begin{equation}
    PSF_L(i,j) = \sum_{k=1}^{n}  A_k*\delta_{L,k}(i,j)  \label{f2}
\end{equation}
where $\delta_{L,k}(i,j)$ represents the energy distribution of ray $A_k$ in the left sub-pixel at DP pixel $(i,j)$.
The ray energy distribution is assumed to be a unit impulse, which equals zero unless it eventually enters the left sub-pixel of DP pixel $(i,j)$.
By following the same calculation steps as for $PSF_L$, $PSF_R$ can also be computed.

\subsection{Pixel-wise DP image rendering module}
As shown in \cref{fig:m1}(b), we reduce the computational cost of ray-traced DP PSFs by training an MLP network to predict them, inspired by~\cite{tseng2021differentiable,yang2023aberration}.
The network input is the normalized coordinate $p$ within the camera's valid imaging region (\cref{fig:m3}(b)), which is a frustum defined by the field of view, sensor size, and the preset minimum and maximum depths. 
We normalize the $(x,y)$ coordinates to $[-1,1]$. 
After fixing the focus distance, we set the DP PSF kernel size ($ks$) large enough to ensure that the DP PSF for any point within the valid imaging region can be fully displayed.
The network consists of 5 hidden layers, each containing 512 neurons, and an output layer with $2*ks^2$ neurons.
We adopt the L2 loss to supervise predictions $\hat{PSF_L}$ and $\hat{PSF_R}$:
\begin{equation}
    Loss = L_2(\hat{PSF_{L}},PSF_{L}) + L_2(\hat{PSF_{R}},PSF_{R})  \label{f5}
\end{equation}

During training, we use ray-traced DP PSFs as GT and apply max normalization to alleviate learning difficulties caused by large radii PSFs.
During inference, we apply sum normalization to the predicted DP PSFs, approximating the uniform intensity distribution in cameras with vignetting compensation.

Because the DP PSF is shift-variant, we convolve each pixel's DP PSF with the AiF image to simulate the DP image captured by a camera.
The RGBD dataset~\cite{silberman2012indoor,mayer2016large} provides the paired RGB image ($I_{RGB}$) and depth map ($I_{D}$).
As shown in \cref{fig:m1}(c), we treat each pixel in $I_{D}$ as an object point and use the trained MLP to predict its DP PSF.
Using $I_{RGB}$ as the AiF image, we apply pixel-wise local convolution~\cite{yang2023aberration} to render the DP image containing aberration and phase information.

\begin{figure*}[t]
    \centering
    \vspace{-1.5mm}
    \includegraphics[width=1\linewidth]{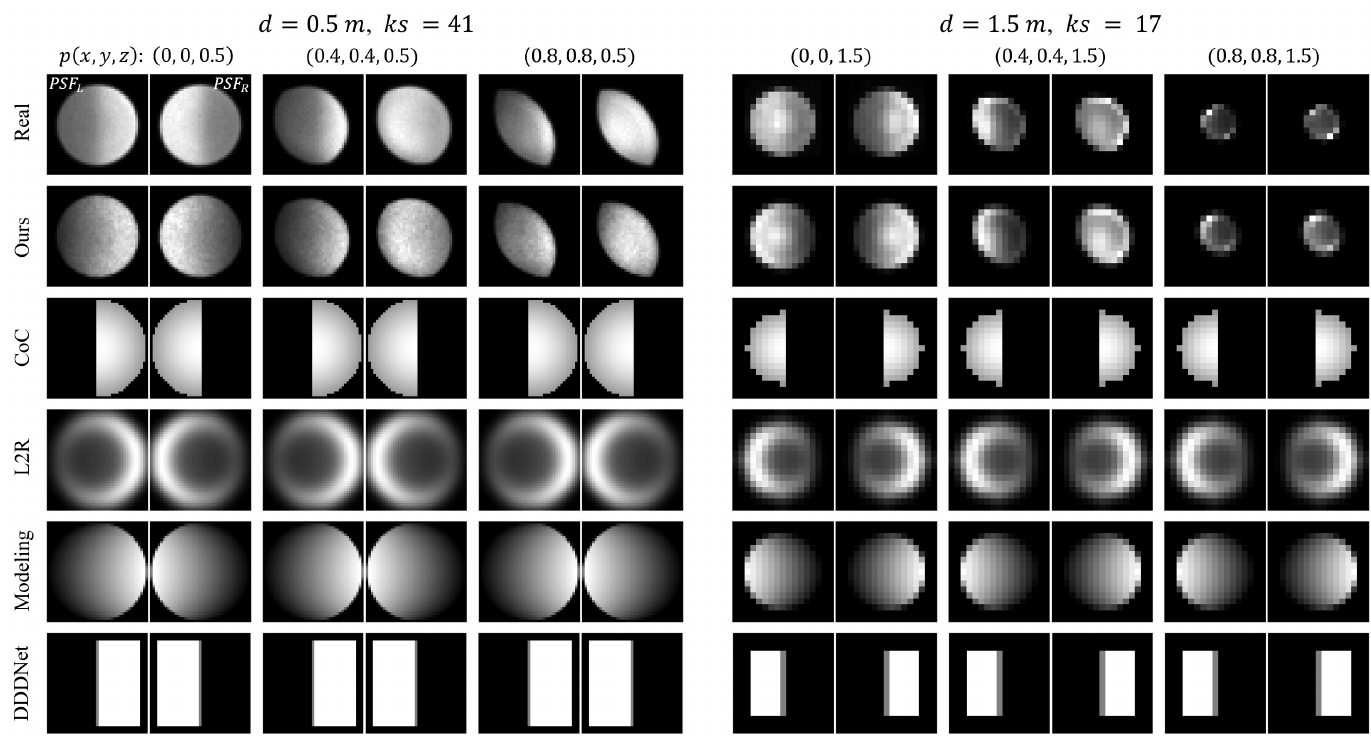}
    \vspace{-7mm}
    \caption{\label{fig:e1}\textbf{Qualitative results of simulated DP PSFs.}
    Evaluate the real and simulated (ours, CoC, L2R~\cite{abuolaim2021learning}, Modeling~\cite{punnappurath2020modeling}, and DDDNet~\cite{pan2021dual}) F/1.8 DP PSFs at two depths (0.5~m and 1.5~m) and three different positions.
    As the object point $p$ moves further from the optical axis, the real $PSF_L$ and $PSF_R$ become more phase asymmetric, and aberrations increase.
    Existing simulators neglect aberrations and DP phase splitting, causing a large gap between simulated and real DP PSFs.
    Only our ray-traced simulator predicts realistic results at all depths and positions.
    }
    \vspace{-3mm}
\end{figure*}

\section{DfDP Model Based on Sdirt}\label{sec:DfDP}
As shown in \cref{fig:m3}(c), the input to the DfDP model is provided in real time by the pixel-wise DP image rendering module of Sdirt.
At each training iteration, Sdirt feeds the DfDP model with simulated DP images of shape $(B, 6, H, W)$,
where $B$ is the batch size, the number of channels is 6,  and $H$, $W$ are the DP image height and width.
The model outputs the predicted depth maps ($\hat{I}_D$) of shape $(B, 1, H, W)$.
We apply the L1 loss to supervise $\hat{I}_D$: 
\begin{equation}
    Loss = L_1(\hat{I_D}, I_D)  \label{f6}
\end{equation}

We select~\cite{cheng2020hierarchical} as the DfDP model and make reasonable adjustments to its cost volume step to accommodate the DP data.
In binocular images, disparity from points at different depths always has the same direction, 
thus the cost volume generation process in~\cite{cheng2020hierarchical} is also unidirectional, as shown by the green arrows in \cref{fig:m3}(d).
However, in DP images, disparity from object points before and after the focus distance is opposite in direction.
Therefore, we add blue arrows to extend the cost volume generation, in order to enhance the DfDP model's ability to capture reverse disparity.

%% file: sec/4_experiment.tex
\section{Sdirt Representation Result}
\label{sec:Sdirt_exp}

\begin{figure*}[t]
    \centering
    \vspace{-1mm}
    \includegraphics[width=1\linewidth]{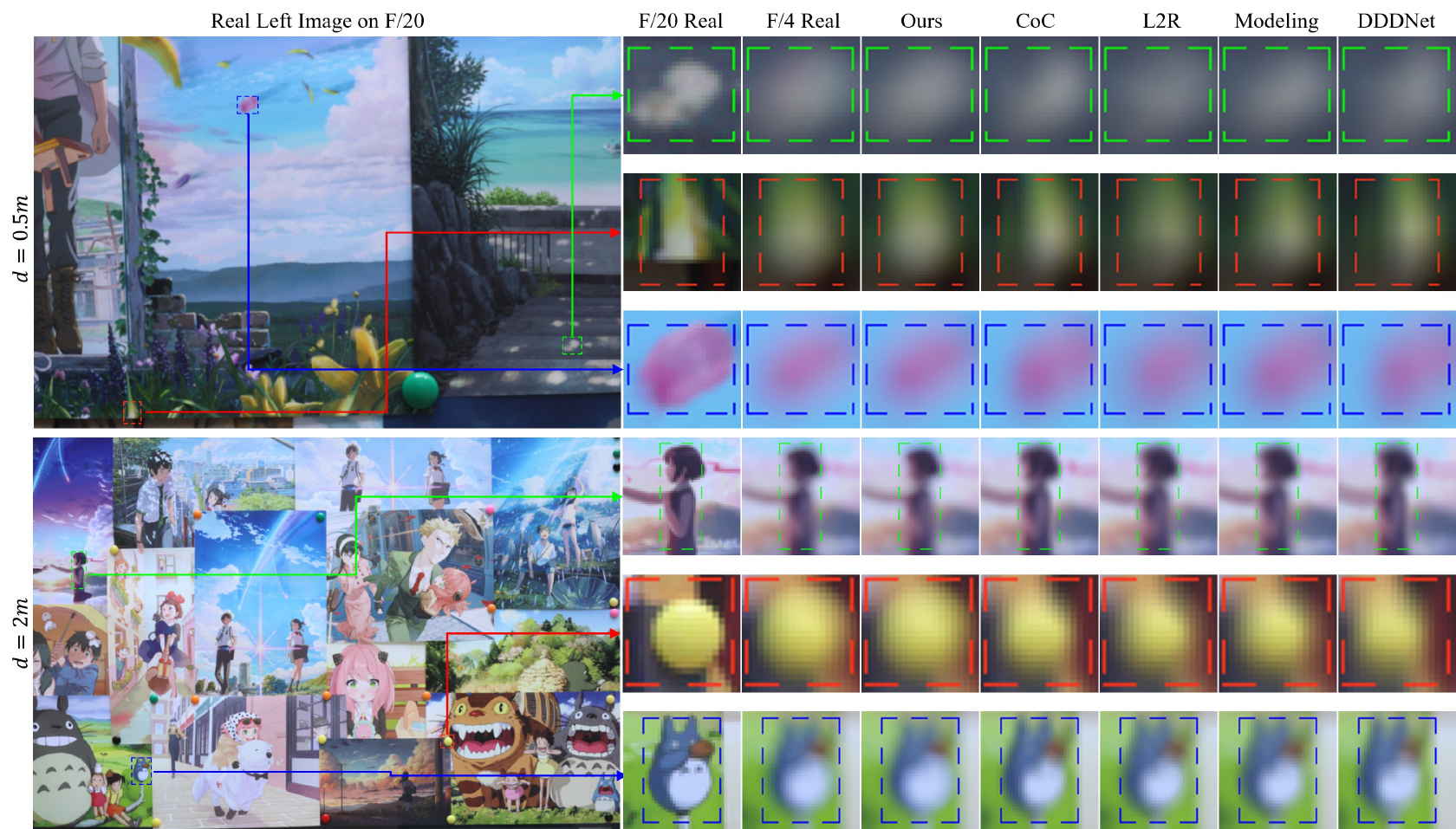}
    \vspace{-7mm}
    \caption{\textbf{Qualitative results of simulated DP images.}
    \label{fig:e2}
    Evaluate the similarity between simulated (ours, CoC, L2R~\cite{abuolaim2021learning}, Modeling~\cite{punnappurath2020modeling}, and DDDNet~\cite{pan2021dual}) and real F/4 defocused left DP images at two depths (0.5~m and 2~m).
    Compared to real F/4 defocused images, images simulated by other methods exhibit varying sizes of patterns, incomplete shapes, and texture shifts in different directions before and after the focus distance (1~m).
    Our method produces the most realistic simulated images.
    }
    \vspace{-3mm}
\end{figure*}

\subsection{Implementation details}
Sdirt employs a Canon RF50mm lens and an R6 Mark II camera body as the real camera.
The simulated DP sensor has dimensions of  $24\,\text{mm} \times 36\,\text{mm}$ with a resolution of $512\times768$. 
We set the focus distance of both the simulated and real cameras to 1~m, the valid imaging range to 0.5--20~m, the F-number to F/4, and the ray set $A$ to contain 4096 rays.
During MLP training, $ks$ is set to 21 to fully display the DP PSF of any point within the valid imaging region. 
We train the MLP for 100,000 iterations, selecting 128 random points within the valid imaging region during each iteration. 
We conducted our experiments on a 12700K CPU and a 3090 GPU.

Following~\cite{yang2023aberration}, we assume chromatic aberration is well corrected compared to other aberrations and defocus effects, 
allowing us to use a 550~nm wavelength to reduce ray tracing costs.
For details on obtaining the structural parameters of the DP pixel, please refer to the Supplementary.

\subsection{Comparison methods}
For comparison, we select all existing model-based DP PSF simulators: 
L2R~\cite{abuolaim2021learning}, Modeling~\cite{punnappurath2020modeling}, DDDNet~\cite{pan2021dual}, and an added CoC simulator.
Although~\cite{pan2024weakly} is a more recent work, its simulator fully reuses DDDNet~\cite{pan2021dual}, so we do not include it for comparison.
Calibration-based simulators and constraint-training methods are data-driven and require real DP-depth paired data, 
making them fundamentally different from model-based simulators, so no comparison is made. 
We directly used the source code of these DP PSF simulators~\cite{punnappurath2020modeling,abuolaim2021learning}, and re-implemented the simulator~\cite{pan2021dual} which only provided executable files.
To ensure fairness, we set the focus distance, F-number, and sensor size in each simulator to be consistent with ours.

\subsection{Evaluation on simulated DP PSFs}

To facilitate observation of aberration and phase information, 
we temporarily set the F-numbers of both the simulated and real cameras to F/1.8. 
We capture a OLED screen displaying a bright pixel to obtain the real DP PSFs. 

As shown in \cref{fig:e1}, we evaluate the DP PSF ($PSF_L$ on the left, $PSF_R$ on the right) of object points $p$ at various depths and positions within the valid imaging region.
Observing the real DP PSFs, it can be found that defocus (with focus distance of 1~m) causes a significant phase difference between the real $PSF_L$ and $PSF_R$, 
and the phase difference at depths of 0.5~m and 1.5~m is opposite.
When $p$ is on the optical axis, all other simulators fail to provide DP PSF results close to the real ones.
Farther from the optical axis, the real $PSF_L$ and $PSF_R$ become phase asymmetric, with smaller phase differences and more off-axis aberrations.
The differences between DP PSFs from other simulators and the real ones become more apparent.
In contrast, our ray-traced simulator predicts accurate results at all depths and positions.

We present the quantitative results in~\cref{tab:e1}, 
using OpenCV's normalized squared difference (NSD) and normalized cross-correlation (NCC) to quantify the error and similarity between the real and simulated DP PSFs.
We sampled 50 DP PSFs at 0.5~m and 1.5~m within the valid imaging region. 
Since the real DP PSF has flip symmetry (not phase symmetry) about both the $x$ and $y$ axes, we performed uniform sampling in the first quadrant.
Our method achieves the highest similarity (0.915 NCC) and lowest error (0.133 NSD), 
demonstrating its effectiveness in bridging the domain gap between simulated and real DP PSFs introduced by lens aberrations and sensor phase splitting.

\begin{table}[h]
    \vspace{-1mm}
    \centering
    \caption{\textbf{Quantitative results of simulated DP PSFs. }
    The NCC and NSD metrics between simulated and real DP PSFs at 50 points were evaluated. Our method achieved the highest similarity and lowest error.}
    \vspace{-3mm}
    \label{tab:e1}
    \resizebox{1\columnwidth}{!}{
    \begin{tabular}{@{}lcccccccc@{}}
      \toprule
      Method            & Ours & CoC & L2R~\cite{abuolaim2021learning} & Modeling~\cite{punnappurath2020modeling} & DDDNet~\cite{pan2021dual}   \\
      \midrule
      NCC$\uparrow$     & \textbf{0.915} & 0.672 & 0.638 & 0.707 & 0.589     \\
      NSD$\downarrow$   & \textbf{0.133} & 0.448 & 0.523 & 0.423 & 0.625     \\
      \bottomrule
    \end{tabular}
    }
    \vspace{-2mm}
\end{table}

In all subsequent experiments, the F-number is adjusted back to F/4 due to the large blur kernel size caused by defocus at F/1.8, which resulted in GPU memory shortages.

\subsection{Evaluation on simulated DP images}
When comparing simulated and real DP images, to assess the degradation of simulated DP images caused by defocus effects at different depths, 
we captured richly textured planar scenes from 0.5~m to infinity, perpendicular to the optical axis. 
Shallower depths were sampled more densely to better capture defocus variations.
This resulted in 56 scenes with varying depths, including 11 infinite-depth scenes. 
For each scene, real DP images were captured at F/4 and F/20. 
The F/20 image served as the AiF RGB image, and its capture-time depth was assigned to the entire depth map.
The F/20 RGB-depth pairs were then input into DP simulators to generate the simulated F/4 defocused DP images.

As shown in \cref{fig:e2}, we compare the qualitative results of the defocused left DP images at depths of 0.5~m and 2~m.
Due to errors in simulating the DP PSF, other methods produce simulated images with textures shifted to the right at 0.5~m depth and to the left at 2~m depth, compared to the real left DP image.
Moreover, other methods show variations in image texture sizes and incomplete shapes.
However, the DP images simulated by our Sdirt method show minimal deviation in texture details.

\begin{figure}[t]
    \centering
    \includegraphics[width=1\linewidth]{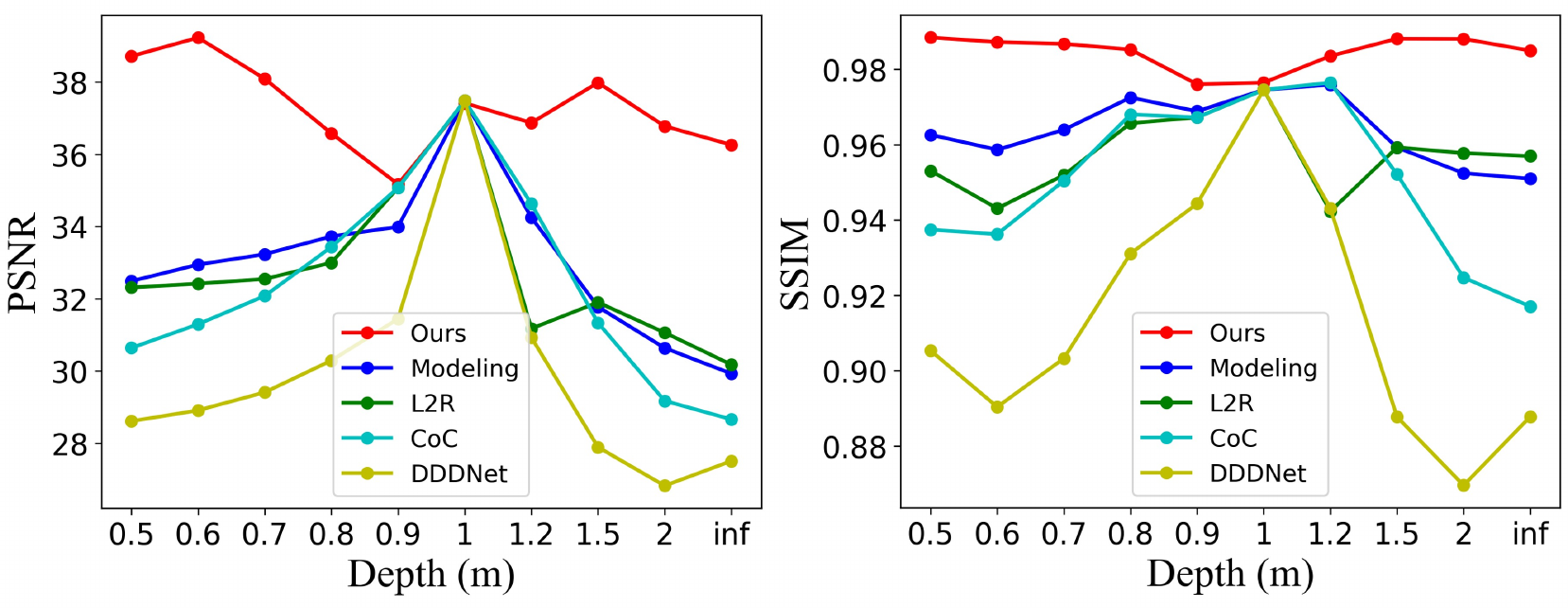}
    \vspace{-8mm}
    \caption{\label{fig:e2_1}
    \textbf{Quantitative results of simulated DP images.}
    Evaluate the similarity of 56 real and simulated (ours, CoC, L2R~\cite{abuolaim2021learning}, Modeling~\cite{punnappurath2020modeling}, and DDDNet~\cite{pan2021dual}) planar scene F/4 DP images at different depths using PSNR ($\uparrow$) and SSIM ($\uparrow$) metrics.
    As the depth of simulated images deviates further from the focus distance (1~m), the less realistic the other methods become, whereas our scheme maintains the highest accuracy across all depths.
    }
    \vspace{-4mm}
\end{figure}

\begin{figure*}[t]
    \centering
    \includegraphics[width=1\linewidth]{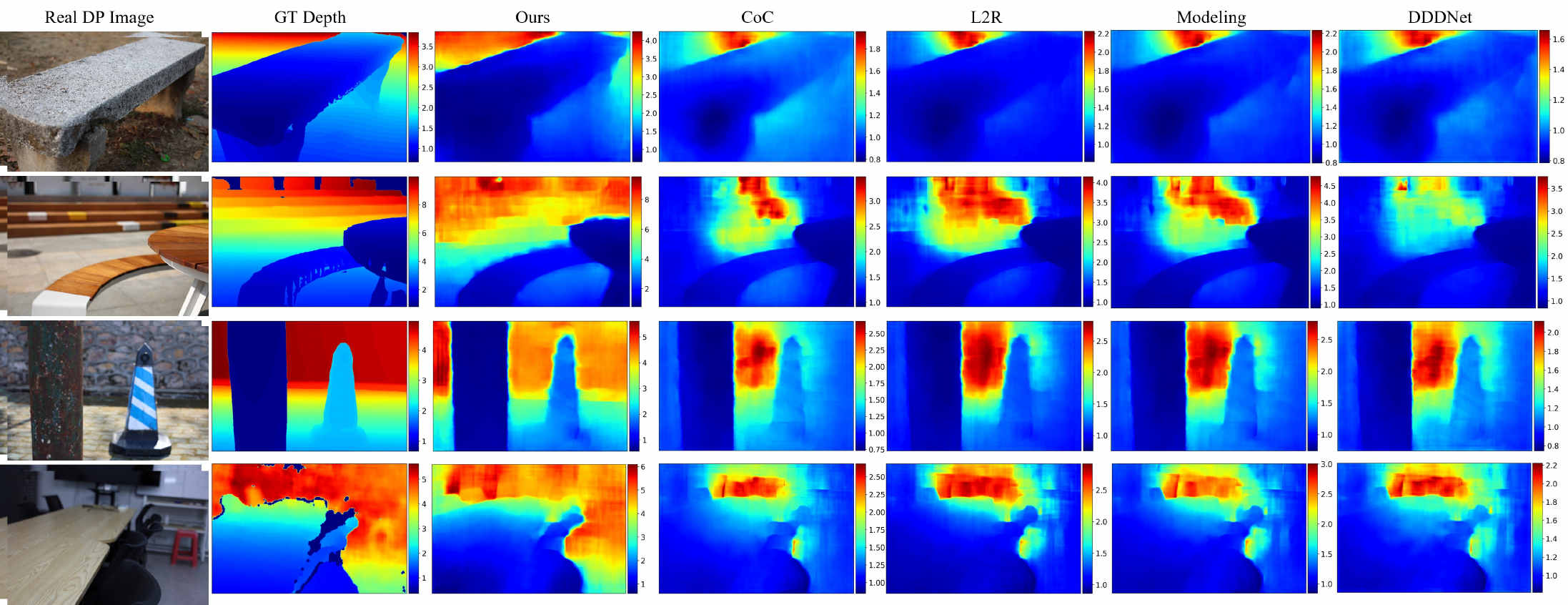}
    \vspace{-7mm}
    \caption{\label{fig:e5}
    \textbf{Qualitative results of absolute depth estimation on DP119.}
    Evaluate DfDP models with CoC, L2R~\cite{abuolaim2021learning}, Modeling~\cite{punnappurath2020modeling}, DDDNet~\cite{pan2021dual} and our Sdirt at four casual scenes.
    Each result image includes a color bar showing depth in meters.
    Their depth estimation results show partial accuracy in relative positional relationships but large absolute positional errors.
    Our depth estimation results, however, demonstrate high accuracy in both relative and absolute positions with minimal errors.
    (Best viewed in color and enlarge on screen.)
    }
    \vspace{-3mm}
\end{figure*}

As shown in \cref{fig:e2_1}, which presents the quantitative results, we assess the similarity between simulated and real images using PSNR and SSIM. 
As the simulated image's depth moves further away from the focus distance (1~m), the neglect of aberrations and phase splitting in other methods leads to a more significant decrease in their metrics.
In contrast, our Sdirt method achieves the best results at all depths, with average PSNR/SSIM values of 37.1982/0.9845, indicating that our simulated DP images are highly realistic.

\section{DfDP Estimation Results}
\label{sec:DfDP_exp}
\subsection{Implementation details}
To train the DfDP model, we simulate F/4 DP images using Sdirt on RGBD datasets~\cite{silberman2012indoor, mayer2016large}. 
The NYU Depth dataset~\cite{silberman2012indoor}, preprocessed by~\cite{Hu2019RevisitingSI}, contains 50,688 indoor scenes. 
We adopt the AdamW optimizer~\cite{loshchilov2017decoupled} and CosineAnnealing scheduler~\cite{loshchilov2016sgdr} with an initial learning rate of $1\times10^{-4}$.
The model is trained with a batch size of 4 for 50 epochs.
In each epoch, 2,000 randomly selected DP-depth pairs are used for training.
After training, the model is directly evaluated on a real-world test set without any fine-tuning to assess its generalization ability.

\subsection{A new real-world test set DP119}
Unlike previous simulation methods, Sdirt addresses data scarcity by simulating hardware, requiring consistent camera and setup during training and inference.
However, as shown in \cref{tab:set1}, no public dataset satisfies all of the following conditions: 
1. Real DP-depth paired data for simulating DP images;
2. A known lens structure for ray tracing;
3. A fixed focus distance, since changing it alters the DP PSF of the same object point.
\,
This makes evaluation on public datasets with unknown hardware impractical.
We therefore collected the DP119 test set, which contains diverse test scenes.

\begin{table}[ht]
    \centering
    \caption{Summary of existing DP datasets.}
    \label{tab:set1}
    \vspace{-3mm}
    \setlength{\tabcolsep}{3pt}
    \resizebox{1\columnwidth}{!}{
    \begin{tabular}{@{}ccccccccc@{}}
      \toprule 
                      & \makecell[c]{DPDD\\ \cite{abuolaim2020defocus}}  &\makecell[c]{L2R\\ \cite{abuolaim2021learning}}  &\makecell[c]{DPNet\\ \cite{garg2019learning}}  &\makecell[c]{DP5K\\ \cite{li2023learning}}  &\makecell[c]{DDDNet\\ \cite{pan2021dual}}   & \makecell[c]{Modeling\\ \cite{punnappurath2020modeling}}  & \makecell[c]{DP119\\Ours }   \\
      \midrule
      Real captured     & $\checkmark$ & $\times$     & $\checkmark$ & $\checkmark$   & $\times$       & $\checkmark$  & $\checkmark$    \\
      Paired depth      & $\times$     & $\checkmark$ & $\checkmark$ & $\checkmark$   & $\checkmark$   & $\times$      & $\checkmark$    \\
      Lens structure    & $\checkmark$ & -            & $\times$     & $\checkmark$   & -              & $\times$      & $\checkmark$     \\
      Fixed focus dist. & $\checkmark$ & -            & $\times$     & $\times$       & -              & $\times$      & $\checkmark$     \\
      \bottomrule
    \end{tabular}
    }
    \vspace{-3mm}
\end{table}

DP119 consists of 45 planar scenes, 44 box scenes, and 30 casual scenes, totaling 119 scenes.
The casual scenes represent common scenarios, suitable for evaluating the robustness of simulation models.
The planar and box scenes are richly textured, helping the model utilize aberration and phase cues for depth estimation.
Textureless areas, even when defocused, lack aberrations or phase differences, making depth estimation difficult and evaluation unreliable.
Thus, richly textured planar and box scenes are ideal for assessing domain gaps between simulated and real DP images.
Real DP images are captured by the Canon RF50mm lens with a R6 Mark II at F/4 and 1~m focus.
GT depth maps for planar scenes are created by the depths at the capture time, while those for box and casual scenes are obtained from LiDAR scans.
For more information about the DP119 test set, please refer to the Supplementary.

\subsection{Evaluation}
We present qualitative and quantitative results for the DfDP model trained by CoC, L2R~\cite{abuolaim2021learning}, Modeling~\cite{punnappurath2020modeling}, DDDNet~\cite{pan2021dual}, and our Sdirt. 
For a fair comparison, we deploy their DfDP models using their DP PSF simulator source code~\cite{abuolaim2021learning, punnappurath2020modeling} and re-implemented code~\cite{pan2021dual}, 
and have all models share the same depth estimation network architecture, initialize from a common CoC pre-trained checkpoint, 
and follow the same implementation details.

\Cref{fig:e5} shows depth estimation results for four casual scenes, with a color bar in meters for visualization.
It can be seen that although other methods provide some relative positional information in the center region, 
they fail to correctly estimate relative positional relationships in the edge regions, and both the center and edge regions have large absolute positional errors.
This is because the DP PSFs predicted by these methods exhibit phase symmetry and shift-invariance, 
whereas real DP PSFs exhibit phase asymmetry, aberrations, and shift-variance.
Their methods' mismatch with real DP PSFs leads to significant depth estimation errors, especially near the edges.
The results of L2R~\cite{abuolaim2021learning} and Modeling~\cite{punnappurath2020modeling} are similar because the peak positions of their DP PSFs are quite close.
In contrast, our method provides accurate relative positional information across the entire image area with minimal absolute positional errors, 
as our simulated DP PSFs are highly consistent with the real ones in both the center and edge regions.

\begin{table}[h]
\centering
\caption{
    \textbf{Quantitative results of absolute depth estimation on DP119.}
    The DfDP model trained with our Sdirt achieves superior performance across most scenarios and evaluation metrics.
}
\vspace{-3mm}
\label{tab:e5}
\resizebox{\columnwidth}{!}{
\begin{tabular}{@{}llllllllll@{}}
    \toprule
    Scene               & Method    & MAE$\downarrow$   & MSE$\downarrow$    & Abs.r.$\downarrow$ & Sq.r.$\downarrow$ & Acc-1$\uparrow$ & Acc-2$\uparrow$  \\
    \midrule
    \multirow{5}{*}{Planar} & Ours    &\textbf{0.0845} & \textbf{0.0109} & \textbf{0.0871} & \textbf{0.0095} & \textbf{0.9849} & \textbf{0.9997}   \\
                        & CoC                                           & 0.2085 & 0.1001 & 0.1801 & 0.0659 & 0.6670 & 0.8990  \\
                        & L2R~\cite{abuolaim2021learning}               & 0.2418 & 0.1271 & 0.2112 & 0.0841 & 0.6319 & 0.8536  \\ 
                        & Modeling~\cite{punnappurath2020modeling}     & 0.2284 & 0.1142 & 0.2004 & 0.0766 & 0.6496 & 0.8725  \\   
                        & DDDNet~\cite{pan2021dual}                     & 0.2583 & 0.1485 & 0.2191 & 0.0958 & 0.5648 & 0.8089  \\
                        \hline
    \multirow{5}{*}{Box} & Ours     & \textbf{0.1197} & \textbf{0.0339} & \textbf{0.0906} & \textbf{0.0231} & \textbf{0.9474} & \textbf{0.9812}  \\
                        & CoC                                           & 0.3375 & 0.1804 & 0.2442 & 0.1116 & 0.4412 & 0.8277  \\
                        & L2R~\cite{abuolaim2021learning}               & 0.3866 & 0.2284 & 0.2803 & 0.1412 & 0.3651 & 0.7156  \\ 
                        & Modeling~\cite{punnappurath2020modeling}     & 0.3655 & 0.2055 & 0.2660 & 0.1278 & 0.3907 & 0.7758  \\     
                        & DDDNet~\cite{pan2021dual}                     & 0.4177 & 0.2676 & 0.2975 & 0.1636 & 0.3456 & 0.6274  \\    
                        \hline         
    \multirow{5}{*}{Casual} & Ours     & \textbf{0.2702} & \textbf{0.2294} & \textbf{0.4632} & 0.7241 & \textbf{0.8236} & \textbf{0.9314}  \\
                        & CoC                                           & 0.7925 & 1.8579 & 0.5461 & 0.6821 &  0.3318 & 0.6103  \\
                        & L2R~\cite{abuolaim2021learning}               & 0.8170 & 1.7487 & 0.5597 & 0.6719 & 0.2760 & 0.5315  \\ 
                        & Modeling~\cite{punnappurath2020modeling}     & 0.7934 & 1.7256 & 0.5510 & \textbf{0.6655} & 0.2978 & 0.5732  \\     
                        & DDDNet~\cite{pan2021dual}                     & 0.8931 & 2.0624 & 0.5752 & 0.7135 & 0.2481 & 0.4685  \\             
\bottomrule
\end{tabular}
}
\vspace{-3mm}
\end{table}

We evaluate all models on the DP119 test set using the following metrics to assess depth estimation performance:
mean absolute error (MAE), mean squared error (MSE),
absolute relative error (Abs.r.), squared relative error (Sq.r.), accuracy with $\delta<1.25$ (Acc-1) and accuracy with $\delta<1.25^2$ (Acc-2).
Since the planar and box scenes are richly textured and do not have interference from textureless areas, 
the simulation models can rely on abundant aberration and phase difference cues for depth estimation. 
As shown in \cref{tab:e5}, in the simplest planar scene, our model achieves the best results across all metrics (0.9849 Acc-1).
In the box scenes, all models experience a performance drop compared to the planar scenes, but we still achieve the optimal metrics (0.9474 Acc-1).
These results demonstrate that our model is highly realistic and effectively bridges the domain gap between simulated and real DP images.
Furthermore, we test the robustness of all models on the casual scenes.
Due to the textureless areas in the casual scenes, the performance of all models significantly degrades. 
However, our model still achieves the best metrics (0.8236 Acc-1), 
far surpassing the second-best model (0.3318 Acc-1).
This further demonstrates that our model is the most realistic and has the best generalization performance.
For more sample results from the DP119 test set, please refer to the Supplementary.

%% file: sec/5_conclu.tex
\section{Conclusion and Discussion}
\label{sec:Conclusion}
Simulated DP images help address the scarcity of DP-depth paired data but face a domain gap between simulated and real DP data. 
In this work, we propose a novel simulation framework called Sdirt to bridge this gap.
Specifically, we calculate the DP PSF for points in object space using ray tracing, and employ a network to predict them.
Then, we render DP images based on the predicted DP PSFs.
Experimental results show that the proposed Sdirt can simulate more realistic DP data.
Moreover, depth estimation models trained based on Sdirt generalize better to real DP images.

We believe that Sdirt is not limited to DfDP tasks, it can provide extra depth information for any task with known optical imaging system parameters, 
promising significant applications in scenarios such as smartphones, automobiles, and microscopes in the future.
\,
However, Sdirt is only applicable to cameras equipped with a fixed-focus lens (the structure must be open) and a DP sensor (DP images must be available).
Currently, only Canon (5D4, R series) meets these requirements. 
To further expand the application of Sdirt, more camera and smartphone manufacturers need to make these data accessible.

\section*{Acknowledgments}
This work was supported by National Natural Science Foundation of China (Grant No. 32471146) and the project N20240194.
The authors thank Echossom, Miya, and Xinge for valuable discussions and assistance.

%% file: sec/X_suppl.tex
\clearpage
\setcounter{page}{1}
\setcounter{section}{0}
\setcounter{figure}{0}
\setcounter{equation}{0}
\setcounter{table}{0}

\renewcommand{\thepage}{S\arabic{page}}
\renewcommand{\thesection}{S\arabic{section}}
\renewcommand{\thesubsection}{\thesection.\arabic{subsection}}
\renewcommand{\thefigure}{S\arabic{figure}}
\renewcommand{\thetable}{S\arabic{table}}
\renewcommand{\theequation}{S\arabic{equation}}

\maketitlesupplementary
\input{sec/0_abstract}
\input{sec/s1_sdirt}

\input{sec/s2_dfdp}


%% file: sec/s1_sdirt.tex
\section{Sdirt Details}
\label{sec:Sdirt_detail}

\begin{figure*}[t]
    \centering
    \includegraphics[width=1\linewidth]{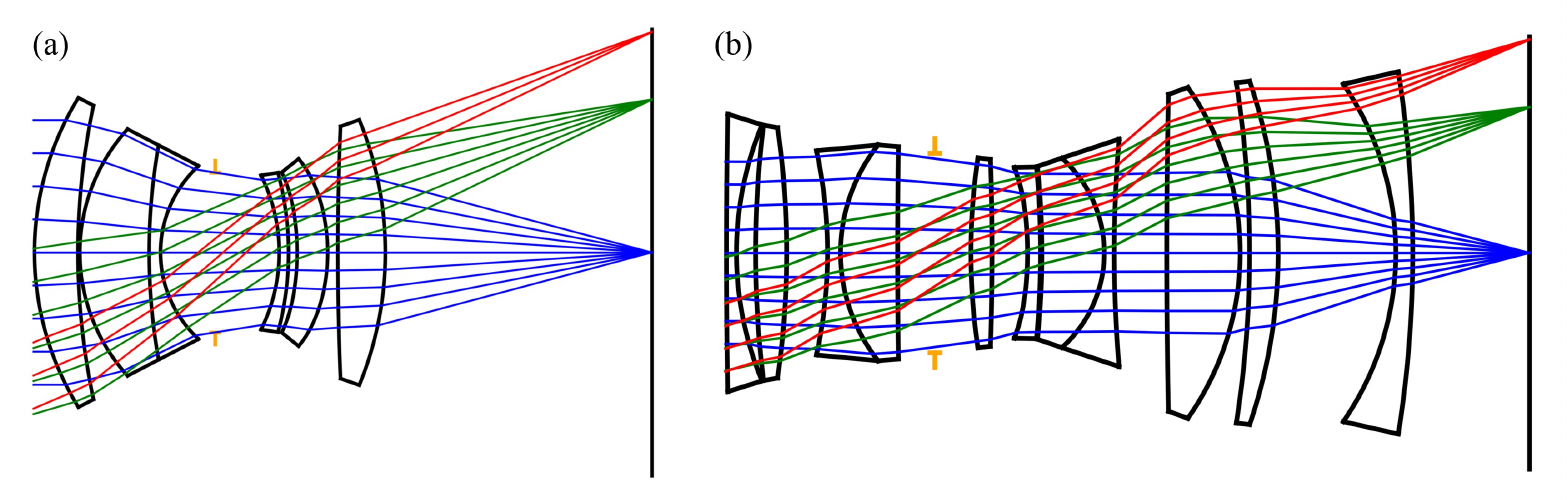}
    \vspace{-6mm}
    \caption{\label{fig:s1}
    \textbf{The 2D lens layout with ray paths.}
    (a) Canon RF50mm F/1.8 lens.
    (b) Canon RF35mm F/1.8 lens.
    }
    \vspace{-3mm}
\end{figure*}

\begin{table*}[t]
    \centering
    \vspace{4mm}
    \caption{
        \textbf{Canon RF50mm F/1.8 lens data.}
        The real reference lens used in the main paper.
    }
    \label{tab:s1}
    \vspace{-2mm}
    \resizebox{\linewidth}{!}{
    \begin{tabular}{@{}lcccccccccccccc@{}}
        \toprule
        Surface    & Radius (mm)  & Thickness (mm)  & Material (n/V) & Diameter (mm) & Conic & $a_4$ & $a_6$ & $a_8$ & $a_{10}$ & $a_{12}$ \\
        \midrule
        1 (Sphere) & 28.621	 & 4.20 &	1.83481/42.7    & 29.99 & 0  & 0 & 0 & 0 & 0 & 0   \\
        2 (Sphere) & 68.136	 & 0.18 &	                & 28.48 & 0  & 0 & 0 & 0 & 0 & 0   \\
        3 (Sphere) & 17.772	 & 6.70 &	1.79952/42.2    & 23.90 & 0  & 0 & 0 & 0 & 0 & 0   \\
        4 (Sphere) & 59.525	 & 1.10 &	1.80518/25.4    & 20.78 & 0  & 0 & 0 & 0 & 0 & 0   \\
        5 (Sphere) & 11.427	 & 5.27 &	                & 16.78 & 0  & 0 & 0 & 0 & 0 & 0   \\
        6 (Aper)   &     	 & 6.20 &	                & 16.24 & 0  & 0 & 0 & 0 & 0 & 0   \\
        7 (Sphere) & -16.726 & 0.90 &	1.67270/32.1    & 14.95 & 0  & 0 & 0 & 0 & 0 & 0   \\
        8 (Sphere) & -29.829 & 0.83 &	                & 15.46 & 0  & 0 & 0 & 0 & 0 & 0   \\
        9 (ASphere) & -25.000 & 2.95 &	1.53110/55.9    & 15.52 & 0  & -4.12032e-05 & -2.90015e-07 & -4.67119e-09 & 7.90646e-11 & -9.28470e-13   \\
        10 (ASphere) & -18.373 & 0.98 &	                & 18.14 & 0  & -2.41619e-05 & -3.29146e-07 & 1.91098e-10  & -9.28593e-13 & -2.29193e-13   \\
        11 (Sphere) & 280.004 & 4.60 &	1.73400/51.5    & 24.43 & 0  & 0 & 0 & 0 & 0 & 0   \\
        12 (Sphere) & -34.002 & 25.67 &	                & 25.71 & 0  & 0 & 0 & 0 & 0 & 0   \\
        Sensor     &          &         &               & 43.27 &    &   &   &   &   &   \\
        \bottomrule
    \end{tabular}
    }
\end{table*}

\begin{table*}[t]
    \centering
    \caption{
        \textbf{Canon RF35mm F/1.8 lens data.}
        The real isolation lens used for detecting DP pixel structural parameters.
    }
    \label{tab:s2}
    \vspace{-2mm}
    \resizebox{\linewidth}{!}{
    \begin{tabular}{@{}lcccccccccccccc@{}}
        \toprule
        Surface    & Radius (mm)  & Thickness (mm)  & Material (n/V) & Diameter (mm) & Conic & $a_4$ & $a_6$ & $a_8$ & $a_{10}$ & $a_{12}$ \\
        \midrule
        1 (Sphere) & 800.000 & 1.00 &	1.80810/22.8    & 27.80 & 0  & 0 & 0 & 0 & 0 & 0   \\
        2 (Sphere) & 33.296	 & 1.92 &	                & 25.64 & 0  & 0 & 0 & 0 & 0 & 0   \\
        3 (Sphere) & 103.801 & 3.11 &	2.00100/29.1    & 25.57 & 0  & 0 & 0 & 0 & 0 & 0   \\
        4 (Sphere) & -86.901 & 4.09 &	                & 25.07 & 0  & 0 & 0 & 0 & 0 & 0   \\
        5 (Sphere) & -47.674 & 1.30 &	1.51742/52.4    & 20.44 & 0  & 0 & 0 & 0 & 0 & 0   \\
        6 (Sphere) & 17.367  & 5.73 &	1.90043/37.4    & 21.60 & 0  & 0 & 0 & 0 & 0 & 0   \\

        7 (Sphere) & 777.674 & 3.72 &	                & 21.26 & 0  & 0 & 0 & 0 & 0 & 0   \\
        8 (Aper)   &     	 & 3.62 &	                & 20.16 & 0  & 0 & 0 & 0 & 0 & 0   \\
        9 (Sphere) & 64.497 & 2.12 &	1.69680/55.5    & 19.04 & 0  & 0 & 0 & 0 & 0 & 0   \\
        10 (Sphere) & -262.934 & 3.56 &	                & 18.69 & 0  & 0 & 0 & 0 & 0 & 0   \\
        11 (ASphere) & -35.963 & 1.30 &	1.58313/59.4    & 17.10 & 0  & -4.61997e-05 & -9.22837e-08 & -4.60687e-10 & 1.65555e-13 & 0   \\
        12 (Sphere) & -93.550 & 0.13 &	                & 17.19 & 0  & 0 & 0 & 0 & 0 & 0   \\
        13 (Sphere) & -84.988 & 6.26 &	1.88300/40.8    & 17.29 & 0  & 0 & 0 & 0 & 0 & 0   \\
        14 (Sphere) & -12.701 & 1.00 &	1.85478/24.8    & 18.97 & 0  & 0 & 0 & 0 & 0 & 0   \\
        15 (Sphere) & 135.000 & 5.27 &	                & 22.77 & 0  & 0 & 0 & 0 & 0 & 0   \\

        16 (Sphere) & 800.000 & 7.35 &	1.90043/37.4    & 31.72 & 0  & 0 & 0 & 0 & 0 & 0   \\
        17 (Sphere) & -28.799 & 0.95 &	                & 33.14 & 0  & 0 & 0 & 0 & 0 & 0   \\
        18 (Sphere) & -109.518 & 2.86 &	1.69680/55.5    & 34.06 & 0  & 0 & 0 & 0 & 0 & 0   \\
        19 (Sphere) & -53.092 & 11.79 &	                & 34.38 & 0  & 0 & 0 & 0 & 0 & 0   \\

        20 (Sphere) & -29.766 & 1.70  &	1.59270/35.3    & 33.78 & 0  & 0 & 0 & 0 & 0 & 0   \\
        21 (Sphere) & -114.300 & 11.66 &	                & 36.27 & 0  & 0 & 0 & 0 & 0 & 0   \\
        Sensor     &          &         &               & 43.27 &    &   &   &   &   &   \\
        \bottomrule
    \end{tabular}
    }
\end{table*}

\begin{figure*}[t]
    \centering
    \includegraphics[width=1\linewidth]{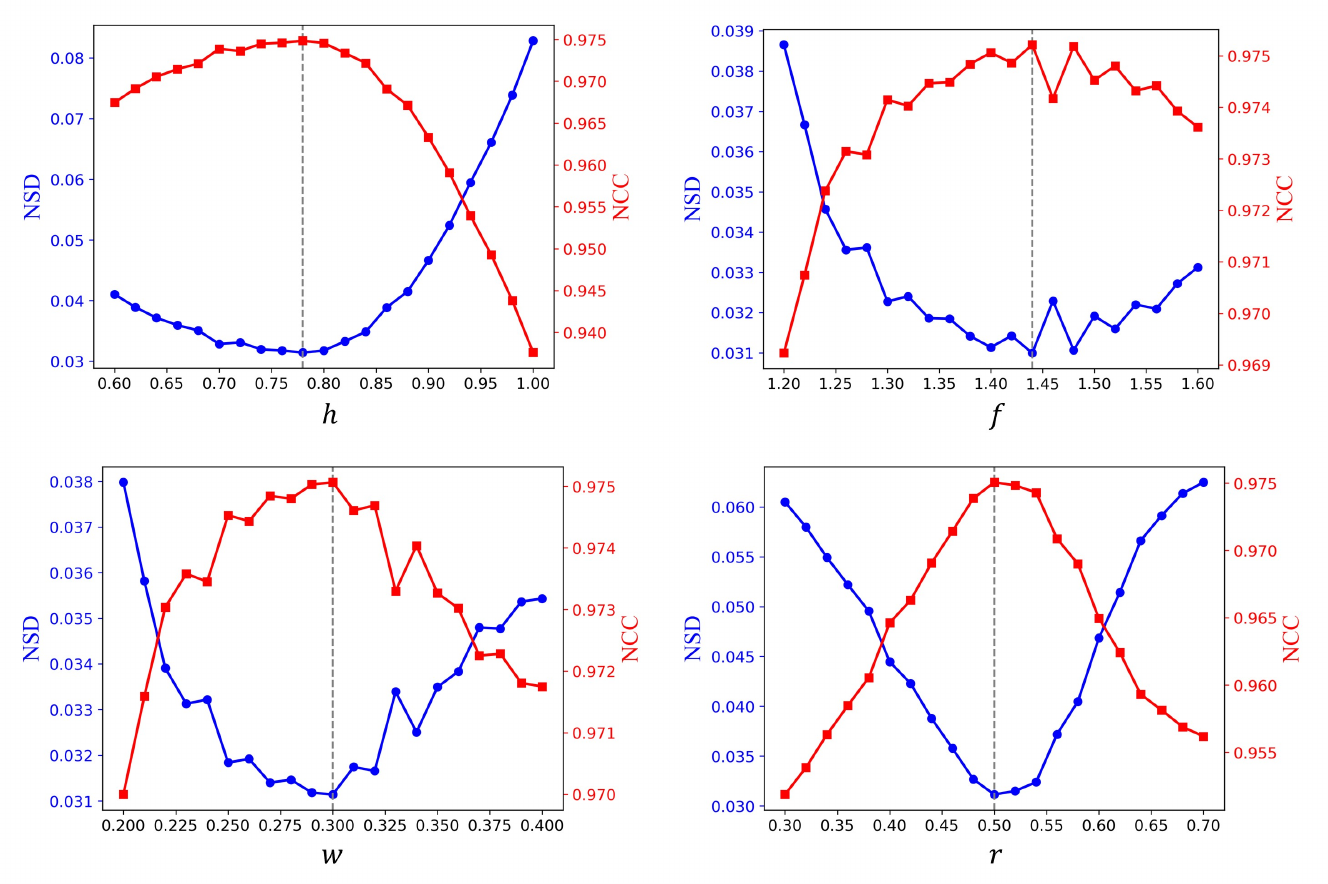}
    \vspace{-8mm}
    \caption{\label{fig:s2}
    \textbf{Error and similarity matching results of DP pixel structural parameters using the isolation lens.}
    For each subplot, the horizontal axis represents the corresponding structural parameter, where the values are given in multiples of the pixel size.
    The left vertical axis represents the NSD matching result between real and simulated DP PSFs, while the right vertical axis represents the NCC matching result.
    To facilitate display, when showing the matching result for any one parameter, we fix the rest of the parameters at their optimal values according to \cref{equ:f1}.
    Moreover, we use white dashed lines to mark the optimal value for each parameter.
    }
    \vspace{-1mm}
\end{figure*}

\subsection{Structural parameters of the DP pixel}
\label{sec:struct}

We select the Canon RF50mm F/1.8 lens and Canon R6 Mark II camera body as our real \textbf{reference lens} and DP sensor in the main paper. 
Since the reference lens data are available, with detailed specifications presented in \cref{tab:s1} and the 2D lens layout with ray paths shown in \cref{fig:s1}(a), 
it is feasible to perform ray tracing on the lens. 
In practice, many commercial lens data, including Canon lenses, are available on open-access websites such as the \href{https://www.photonstophotos.net/GeneralTopics/Lenses/OpticalBench/OpticalBench.htm}{Optical Bench}, which facilitates optical modeling.  
However, performing ray tracing on the DP sensor is limited by camera manufacturers' nondisclosure of the structural parameters of the microlens and sub-pixel components within the DP pixel.
Therefore, as described in Sec. 3.1 (Ray-traced DP PSF simulator) and Fig. 3(a) (DP pixel structure layout) of the main paper, 
we simplify the DP pixel structure by modeling the microlens as a thin lens with radius $r$ and focal length $f$,
defining $h$ as the distance between the sub-pixel and the microlens, $w$ as the sub-pixel width, and $ps$ as the DP pixel size.

We assign a set of possible values to each structural parameter of the DP pixel and identify the optimal parameter combination through grid searching across these parameters.
Instead of conducting search experiments directly on the reference lens,
we employ the Canon RF35mm F/1.8 as an \textbf{isolation lens} to replace the reference lens for probing DP pixel structural parameters, 
thereby ensuring isolation between the reference lens and the sensor.
Detailed lens data for the Canon RF35mm F/1.8 are provided in \cref{tab:s2}, and its 2D lens layout with ray paths is shown in \cref{fig:s1}(b).
In the valid imaging region (Fig. 4(b) in the main paper), we select 5 object points at different positions and capture real DP PSFs through the employed camera.
By evaluating the error and similarity between the real DP PSFs and the set of DP PSFs simulated with all DP pixel structural parameters,
we determine an optimal parameter combination:
\begin{align}
\begin{cases}
    h &= 0.78*ps \\
    f &= 1.44*ps \\
    w &= 0.30*ps \\
    r &= 0.50*ps \label{equ:f1}
\end{cases}
\end{align}

Specifically, we choose the normalized squared difference (NSD) and normalized cross-correlation (NCC) methods
from OpenCV to evaluate the error and similarity between the real and simulated DP PSFs.
During the search experiments, we set the F-number to F/4.0, set the DP PSF kernel size to 35, and set the focal distance to infinity.
As shown in \cref{fig:s2}, we present the results of the error and similarity matching.
For ease of presentation, when displaying the matching results for any one parameter, we fix the other parameters at their optimal values according to \cref{equ:f1}.

\begin{figure*}[t]
    \centering
    \vspace{3mm}
    \includegraphics[width=1\linewidth]{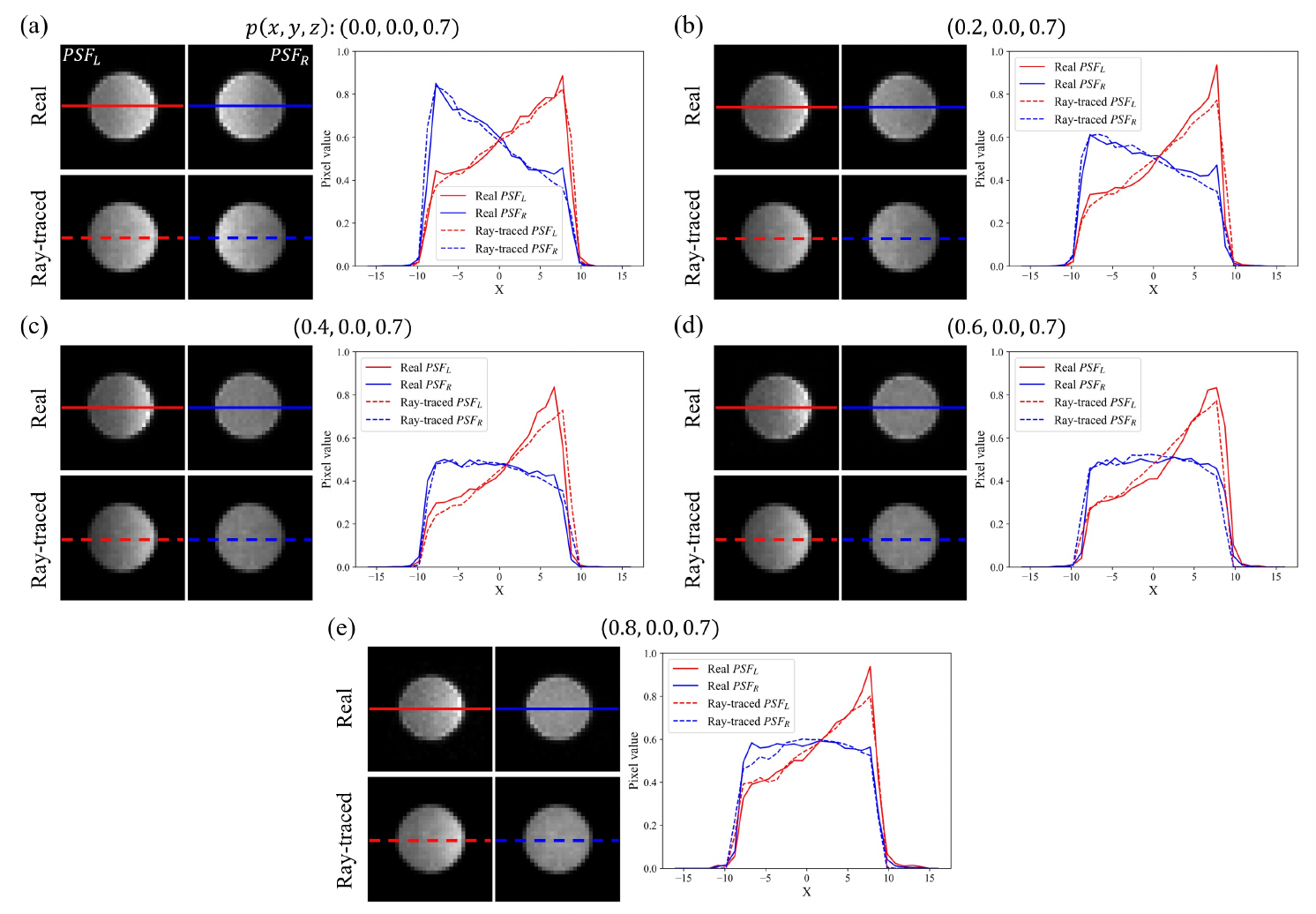}
    \vspace{-7mm}
    \caption{\label{fig:s3}
    \textbf{Qualitative results of DP PSF for real and ray tracing.}
    We select 5 points within the valid imaging region, and only their x-coordinates increased sequentially, corresponding to (a) - (e).
    We provide not only the comparison results of real and ray-traced DP PSF ($PSF_L$ on the left, $PSF_R$ on the right) for these 5 points.
    We also present the pixel distribution curves for the central row.
    As the object point $p$ moves farther away from the optical axis, the real $PSF_L$ and $PSF_R$ become more phase asymmetric.
    Our simulated DP PSFs, obtained through ray tracing, align well with the real DP PSFs.
    }
\end{figure*}

To demonstrate the similarity between real and simulated DP PSFs under the optimal combination of structural parameters, 
we conduct a qualitative analysis while keeping the lens, F-number, kernel size, and focal distance unchanged from the grid search experiments.
As shown in \cref{fig:s3}, we select five points evenly spaced within the valid imaging region, with only their x-coordinates increasing sequentially.
Additionally, to visually compare changes in pixel values, a line plot of pixel distribution along the central row is also included in each subplot.
Observing the actual DP PSF, we notice that as the object point $p$ moves farther from the optical axis, the real $PSF_L$ and $PSF_R$ become more phase asymmetric.
This observation aligns with the experimental results presented in Sec. 5.3 (Evaluation on simulated DP PSFs) of the main paper.
When comparing the ray-traced and real DP PSFs, we find that the ray-traced DP PSF is not only highly realistic in morphology but also closely matches the real data in pixel values.

\subsection{DP PSF capture and linearization}
The native resolution of the Canon R6 Mark II is $6000 \times 4000$ pixels. Due to GPU memory limitations, we reduced the spatial resolution to $768 \times 512$.
Although this resolution is far below the original, our experiments on DP PSFs (see Fig. 5 in the main paper and \cref{fig:s3}) demonstrate that aberrations and phase differences in defocused regions remain clearly visible.

Capturing the real DP PSFs is a straightforward process.
A single white pixel is displayed on an OLED screen (iPhone 15 Pro, 460 ppi, 55~um pixel size).
We capture the PSFs using a Canon R6 Mark II (6~um pixel size) equipped with an RF50mm lens at object distances ranging from 0.5~m to 2~m.
For the camera's native resolution of $6000 \times 4000$, the Nyquist-limited spatial resolution is approximately 120~um at 0.5~m and 480~um at 2~m.
Given the subsequent downsampling, capturing a single bright pixel on the screen is sufficient to ensure the acquisition of the real PSFs.

\begin{figure*}[t]
    \centering
    \vspace{10mm}
    \includegraphics[width=1\linewidth]{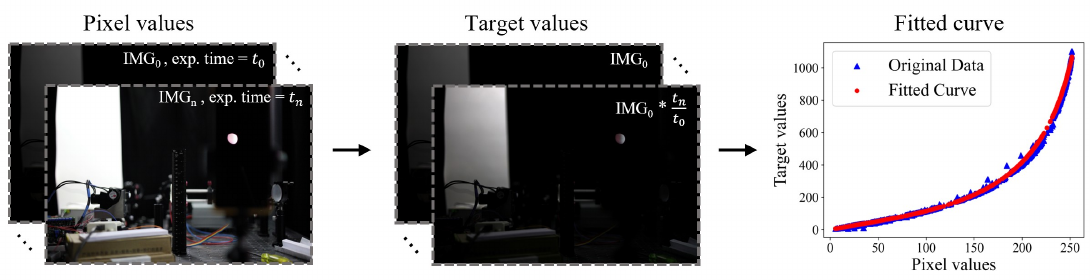}
    \vspace{-7mm}
    \caption{\label{fig:r1}
    \textbf{Overview of our linear calibration method.} 
    The plot shows three components: pixel values from images captured under varying exposure times, 
    target values derived by scaling the shortest exposure image (pixel values within [0, 35]) according to exposure time ratios, 
    and the fitted curve obtained by applying a piecewise weighted fitting approach to align original pixel values with target values. 
    The fitted functions are publicly available in our released code.
    }
\end{figure*}

The PSFs generated by optical simulations represent linear light intensity distributions, similar to those in the camera's RAW domain.
To our knowledge, Canon's RAW-domain dual-pixel images are not publicly available, and currently only RGB-domain dual-pixel images can be obtained.
Therefore, the PSFs captured by the camera are pixel value distributions in the RGB domain, 
transformed from RAW-domain light intensities by the image signal processing (ISP) pipeline.
This pipeline involves non-linear and non-invertible operations such as gamma correction and contrast enhancement.
Although modern ISP pipelines are highly complex, 
radiometric calibration remains a common and practical approach to approximately linearize RGB data and recover the underlying light intensity distributions in the RAW domain.
As shown in \cref{fig:r1}, our linear calibration method consists of:
1. Capture a static scene under varying exposure times;
2. Use the image captured with the shortest exposure time (pixel values within [0, 35]) as a linear reference,
scaling it according to exposure time ratios to obtain target values for the other images;
3. Fit the original pixel values of other images to these targets. 
We adopt a piecewise weighted fitting approach, and the fitted functions are released with our code.

\subsection{Local DP PSF convolution}
Existing methods generally use the same PSF kernel to convolve the entire image.
Yang \etal~\cite{yang2023aberration} provides a local convolution operation function based on PyTorch, enabling the use of different PSF kernels for each pixel.
However, in our experiments, we use different DP PSF kernels for each pixel.
We corrected the kernel flipping error in his function and provided an implementation for using different DP PSF kernels for each pixel, as follows:
\lstset{language=python} 
\begin{lstlisting}[breaklines]
def local_dp_psf_render(input, dp_psf, kernel_size=21):
    """ Render DP image with local DP PSF. 
    Use different DP PSFs for different pixels.
    Args:
        input (Tensor): The image to be blurred (N, C, H, W).
        dp_psf (Tensor): Per pixel local DP PSFs (1, H, W, 2, ks, ks)
        kernel_size (int): Size of the DP PSFs. Defaults to 21.
    Returns:
        output (Tensor): Rendered DP image (N, 2*C, H, W)
    """

    b,c,h,w = input.shape
    pad = int((kernel_size-1)/2)

    # 1. pad the input with replicated values
    inp_pad = torch.nn.functional.pad(input, pad=(pad,pad,pad,pad), mode='replicate')

    # 2. Create a Tensor of varying DP PSF
    kernels = dp_psf.reshape(-1, 2, kernel_size, kernel_size)
    kernels_flip = torch.flip(kernels, [-2, -1])
    kernels_rgb = torch.stack(c*[kernels_flip], 2)

    # 3. Unfold input
    inp_unf = torch.nn.functional.unfold(inp_pad, (kernel_size,kernel_size))   
    
    # 4. Multiply kernel with unfolded
    x1 = inp_unf.view(b,c,-1,h*w)
    x2_l = kernels_rgb[:,0,...].view(b, h*w, c, -1).permute(0, 2, 3, 1)
    x2_r = kernels_rgb[:,1,...].view(b, h*w, c, -1).permute(0, 2, 3, 1)
    y_l = (x1*x2_l).sum(2)
    y_r = (x1*x2_r).sum(2)
    
    # 5. Fold output
    render_l = torch.nn.functional.fold(y_l,(h,w),(1,1))
    render_r = torch.nn.functional.fold(y_r,(h,w),(1,1))

    return torch.cat([render_l,render_r], dim=1)
\end{lstlisting}

\subsection{DP PSF prediction network}

\begin{figure}[t]
    \centering
    \includegraphics[width=1\linewidth]{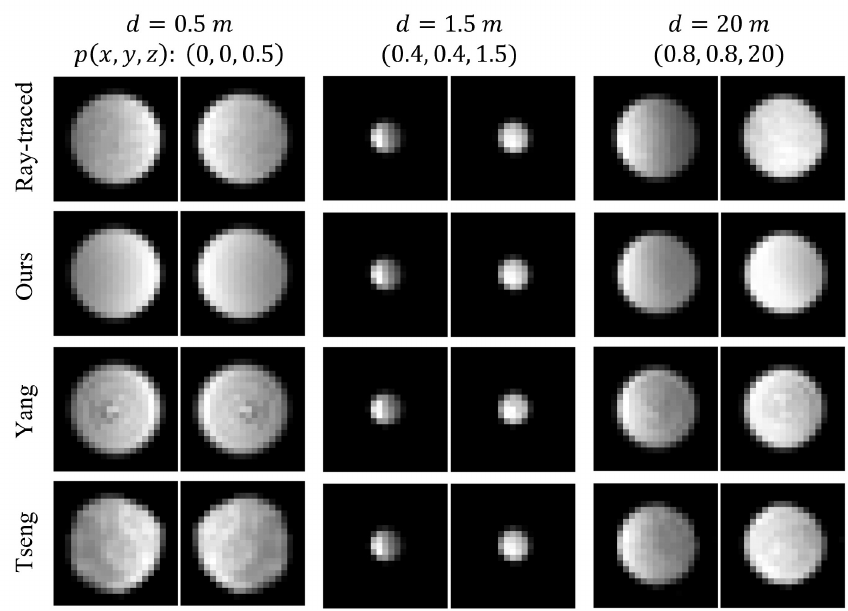}
    \vspace{-7mm}
    \caption{\label{fig:e2}\textbf{Qualitative results of DP PSF prediction networks.}
    Evaluate the ray-traced and network-predicted (ours, Yang \etal~\cite{yang2023aberration}, and Tseng \etal~\cite{tseng2021differentiable}) F/4 DP PSFs at three depths (0.5m, 1.5m, 20m) and positions.
    For small DP PSF radii, all network predictions closely match ray-traced results. 
    For larger radii, the predicted results by~\cite{yang2023aberration,tseng2021differentiable} show significant deviations, 
    attributed to differences in normalization schemes with ours during training.
    }
    \vspace{-2mm}
\end{figure}

During the training of the DP PSF prediction network, as described in Sec. 5.1 (Implementation details) of the main paper, we set the aperture to F/4, ks to 21, and trained for 100,000 iterations.
After training, we evaluated the fitting quality of the DP PSF network.

We present qualitative and quantitative comparison results between our MLP network and recent work~\cite{yang2023aberration,tseng2021differentiable},
As shown in \cref{fig:e2}, we evaluate DP PSF at three depths and positions, using ray-traced DP PSF as GT.
Their results show significant errors compared to ray-traced ones at large PSF radii.
This is due to their training scheme using sum normalization for GT, which is effective for small PSF radii but becomes less accurate for larger ones.
In contrast, our network, also using only MLP, can predict accurate results at all depths and positions, even at large PSF radii.
The only difference is that we use max normalization for GT DP PSFs during training, and apply sum normalization to the predicted DP PSFs during inference to approximate the uniform intensity distribution of cameras with vignetting compensation.
Compared to the DP PSF obtained through ray tracing, the results output by the MLP network are smoother. 
This is because the number of rays sampled in ray tracing is limited by memory and computation time, with only 4096 rays set. 

\begin{table}[h]
    \centering
    \caption{Quantitative results of DP PSF prediction networks.}
    \vspace{-3mm}
    \label{tab:e2}
    \resizebox{0.8\columnwidth}{!}{
    \begin{tabular}{@{}llll@{}}
      \toprule
      Method    & L1 error $\downarrow$ & L2 error $\downarrow$ & Time (s) $\downarrow$    \\
      \midrule
      Ray-traced                                    &  -          &    -      & 801.544     \\
      Ours               & \textbf{6.887e-5}    & \textbf{6.830e-8}  & \textbf{0.395 }      \\
      Yang \etal~\cite{yang2023aberration}          & 1.487e-4    & 1.381e-7  & \textbf{0.395}  \\
      Tseng \etal~\cite{tseng2021differentiable}    & 1.124e-4    & 7.685e-7  & 43.96       \\
      \bottomrule
    \end{tabular}
    }
\end{table}

We present quantitative results in~\cref{tab:e2}. Specifically, we compare L1 and L2 errors for 50 different depths and positions, as well as the time cost for predicting a DP PSF map for all pixels (a total of $512\times768$) in the depth map.
Our network outperforms those developed by Yang \etal~\cite{yang2023aberration} and Tseng \etal~\cite{tseng2021differentiable} in accuracy, 
and its low time cost enables high-speed image rendering.
In contrast, the network proposed by Tseng \etal~\cite{tseng2021differentiable} incurs high time costs due to its inclusion of convolutional layers.

Overall, rendering a DP image using a DP PSF map is very fast (with a fixed cost of 0.2 seconds).
However, generating these DP PSFs through ray tracing is time-consuming (801.544 seconds).
We use MLP to save time, achieving high speed (0.395 seconds), which enables real-time rendering of DP images for DP data-driven tasks during each iteration.

%% file: sec/s2_dfdp.tex
\section{Depth-from-Dual-Pixel Details}
\label{sec:DfDP_detail}
\subsection{Adjustment of cost volume}
In the field of DfDP, we do not pursue the optimal DfDP network structure. 
Instead, we select~\cite{cheng2020hierarchical} as the DfDP model and make reasonable adjustments to its cost volume step to accommodate the DP data.

Existing binocular disparity matching algorithms typically only consider unidirectional disparity shifts when stacking left and right image features to form the cost volume.
This is because in binocular images, the disparity formed by points at any depth is always in the same direction.
However, in DP images, the disparity direction formed by object points before and after the focal distance is opposite.
Therefore, as shown in Fig. 4(d) (main paper), we add reverse disparity while stacking the original disparity.
Our implementation is as follows:
\lstset{language=python} 
\begin{lstlisting}[breaklines]
def get_dp_cost_volume(x, y, d_max=20):
    """ Get DP image cost volume 
    Stack original disparity and add reverse disparity.
    Args:
        x (Tensor): Left DP image feature (B, C, H, W).
        y (Tensor): Right DP image feature (B, C, H, W).
        d_max (int): Max displacement. Defaults to 20.
    Returns:
        cost (Tensor): Cost volume of DP image features (B, 2*C, d_max, H, W)
    """

    B, C, H, W = x.size()
    cost = torch.zeros(B, C*2, d_max, H, W).type_as(x)
    for i in range(d_max):
        d = i-d_max//2
        if d < 0:
            cost[:, :C, i, :, :d]=x[:, :, :, :d]
            cost[:, C:, i, :, :d]=y[:, :, :, -d:]
        elif d == 0:
            cost[:, :C, i, :, :]=x
            cost[:, C:, i, :, :]=y
        if d > 0:
            cost[:, :C, i, :, d:]=x[:, :, :, d:]
            cost[:, C:, i, :, d:]=y[:, :, :, :-d]

    return cost
\end{lstlisting}

\subsection{A new real-world test set DP119}

We used the Canon RF50mm lens and Canon R6 Mark II camera to capture a new dataset, setting the aperture to F/4 and focusing at 1~m. 
Each scene provides real DP-depth paired data. 
Finally, we collected the DP119 dataset, which includes 45 planar scenes, 44 box scenes, and 30 casual scenes, totaling 119 scenes.

\begin{figure*}[t]
    \centering
    \includegraphics[width=1\linewidth]{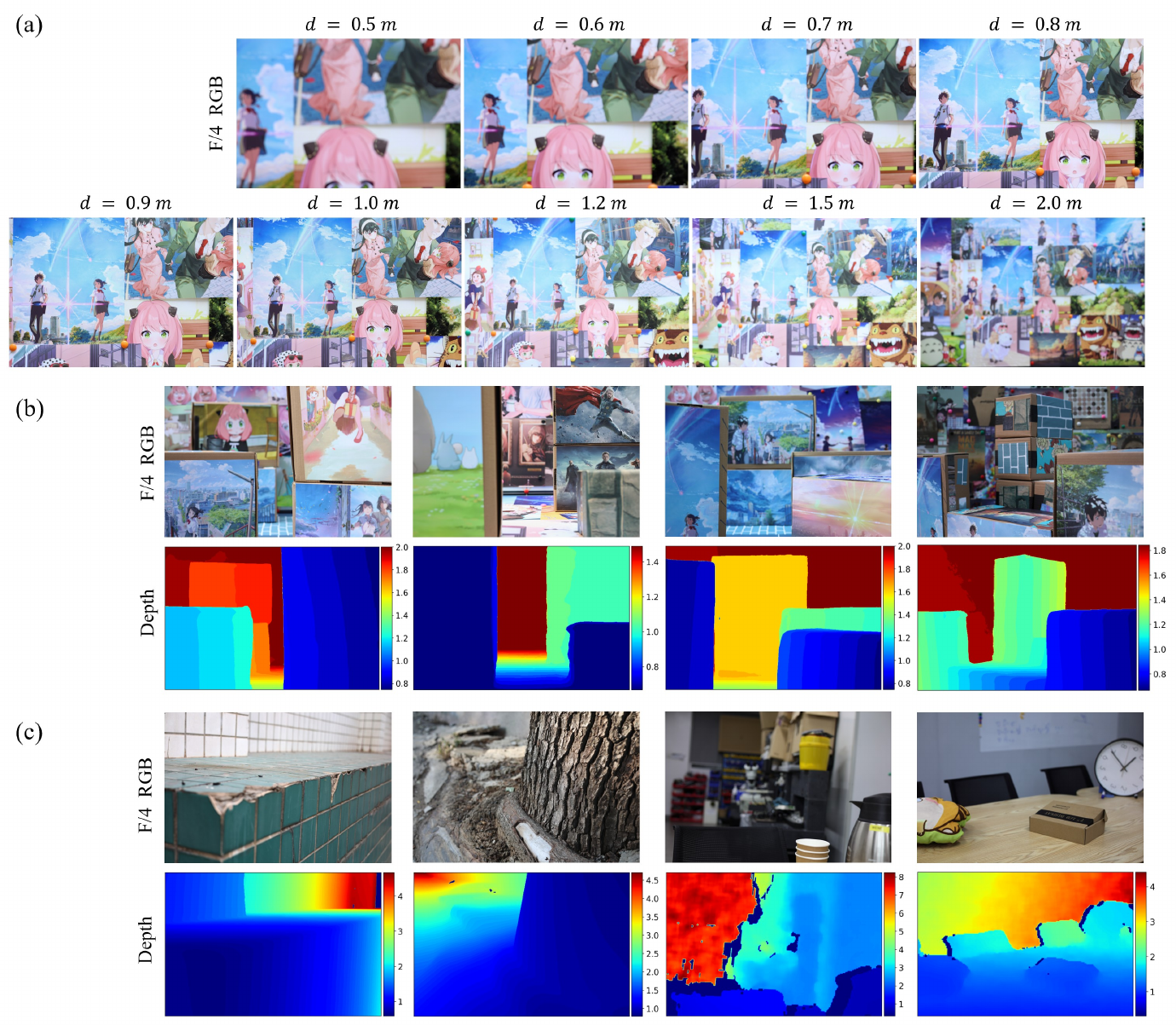}
    \vspace{-7mm}
    \caption{\label{fig:s5}
    \textbf{Examples of scenes in the DP119 dataset.} Each depth map is decorated with a color bar in meters.
    (a) Planar scene samples. The $d$ is the planar depth value at the time of capture. (b) Box scene samples. (c) Casual scene samples. 
    }
    \vspace{-2mm}
\end{figure*}

For the planar scenes, we captured richly textured posters images from 0.5~m to 2~m perpendicular to the camera's optical axis, with denser sampling at shallower depths.
The dataset includes five distinct scenes at nine different distances (0.5~m, 0.6~m, 0.7~m, 0.8~m, 0.9~m, 1.0~m, 1.2~m, 1.5~m, 2~m).
Real DP images were captured at apertures F/4 and F/20, adjusting ISO to match light intensity. 
F/20 images were used as AiF images, with their depth value copied into the depth map.
Unlike box and casual scenes, planar scenes have simple depth structures, which avoid common issues such as holes in LiDAR depth maps and misalignment between depth and RGB images.
These scenes are useful for evaluating both depth estimation models and the realism of DP simulators.
As shown in \cref{fig:s5}(a), we present a sample set of planar scenes.

For the box scenes, we covered all the boxes and backgrounds with posters.
The boxes were randomly placed within a range of 0.5~m to 2~m, and each time we took a shot, 
we significantly adjusted the number, texture, and placement of the boxes, deliberately creating situations of overlap, tilt, and occlusion.
Depth was captured using the LiDAR of an iPhone 15 Pro.
Both planar and box scenes are richly textured, aiding the model in utilizing aberration and phase difference cues for depth estimation.
Textureless areas, even when defocused, do not introduce any aberrations or phase differences, which would interfere with depth estimation.
Therefore, planar and box scenes are ideal for evaluating whether the DP simulator addresses the domain gap between simulated and real DP images.
As shown in \cref{fig:s5}(b), we present samples of box scenes.

For the casual scenes, we directly captured ordinary real-world scenes, intentionally controlling the depth within 0.5~m to 10~m, 
as scenes beyond 10m would yield unreliable results from the depth sensor.
Among the 30 casual scenes, 20 indoor scenes were captured with depth obtained using a binocular structured-light camera (Orbbec Gemini2). 
For the 10 outdoor scenes, since the binocular structured-light camera performs poorly outdoors, we used the LiDAR of an iPhone 15 Pro to capture depth.
As shown in \cref{fig:s5}(c), we present samples of casual scenes.

\subsection{More DfDP results on the DP119 test set}

\begin{figure*}[t]
    \centering
    \includegraphics[width=1\linewidth]{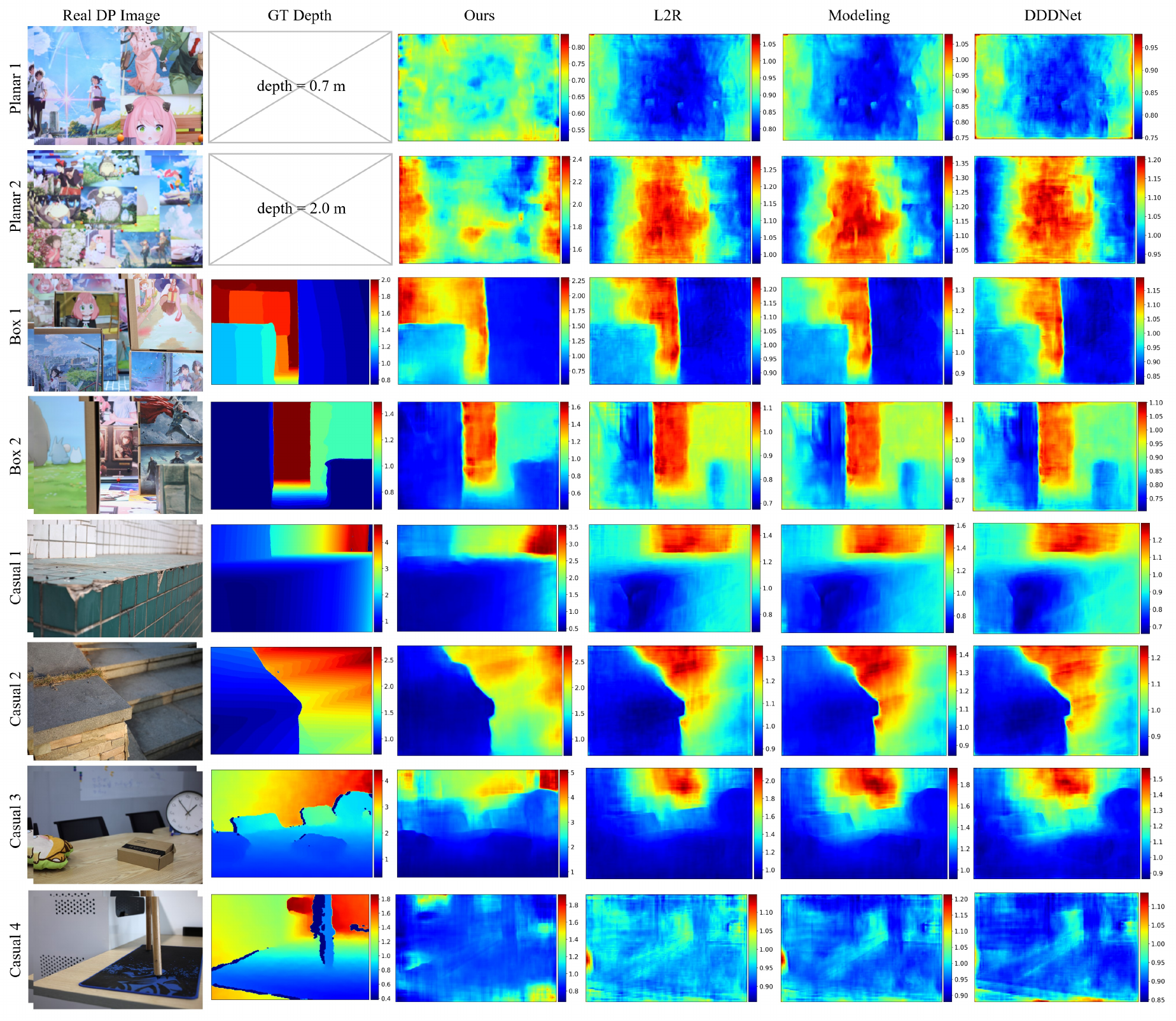}
    \vspace{-7mm}
    \caption{\label{fig:s6}
    \textbf{The depth estimation results from different simulators on the DP119 test set.}     
    Evaluate DfDP models with L2R~\cite{abuolaim2021learning}, Modeling~\cite{punnappurath2020modeling}, DDDNet~\cite{pan2021dual} and our Sdirt on various DP119 dataset scenes.
    Each result image is decorated with a color bar in meters.
    Their depth estimation results show partial accuracy in relative positional relationships but large absolute positional errors.
    Our depth estimation results, however, demonstrate accurate in both relative and absolute positions with minimal errors.
    Furthermore, textureless areas lead to degradation in all models.
    (Best viewed in color and enlarge on screen.)
    }
    \vspace{6mm}
\end{figure*}

\cref{fig:s6} presents the depth estimation results of various simulators on different scenes in the DP119 dataset.
From planar scenes 1 and 2, we observe that due to increased aberrations and asymmetric phase differences at the edge regions,
all models, including ours, exhibit significant relative positional errors in both central and edge regions. 
However, our model provides more accurate absolute depth estimates, whereas other models tend to bias toward the focus distance (1~m). 
The depth estimation results of L2R~\cite{abuolaim2021learning} and Modeling~\cite{punnappurath2020modeling} are similar, as their DP PSF peak positions are spatially close.
A similar trend is observed in box scenes 1 and 2, where other models show a strong bias toward the focal distance (1~m) and exhibit large relative positional errors in both central and edge regions. 
In contrast, our model maintains superior performance.

In casual scenes 1 and 2, the textures are rich, and the scenes are simple. 
Models can predict depth not only using aberrations and phase differences but also by leveraging additional structural cues.
As a result, the relative positional errors in the center and edge regions were alleviated for all models in these two scenes. 
However, the absolute depth estimates from other models still show a clear bias toward the focus distance (1~m).
Moreover, in casual scenes 3 and 4, large textureless areas appear, missing aberrations and phase differences. 
Depth estimation models can only infer the depth of these textureless areas along structural cues, 
leading to significant relative positional errors in textureless regions for all models.

Overall, our model outperforms others in both relative and absolute position accuracy. 
This demonstrates the high realism of our DP simulator and its ability to bridge the domain gap between simulated and real DP images.